\documentclass[runningheads]{llncs}

\usepackage{eccv}

\usepackage{xcolor}
\usepackage{booktabs}
\usepackage{makecell}
\usepackage{cuted}
\usepackage{microtype}
\usepackage{bm}
\usepackage{subcaption} 
\definecolor{eccvblue}{rgb}{0.21,0.49,0.74}
\usepackage[breaklinks,colorlinks,allcolors=eccvblue]{hyperref}

\def\confName{ECCV}
\def\confYear{2026}
\def\ours{MCGC\xspace}
\newcommand{\dataseturl}{\url{https://github.com/MinweiZhao/GBA-GCs}}

\usepackage{eccvabbrv}

\usepackage{graphicx}
\usepackage{booktabs}

\usepackage[accsupp]{axessibility}  

\usepackage{orcidlink}

\begin{document}

\title{Urban Boundaries, Social Barriers: A Benchmark and Vision-Centric Framework for Mapping Gated Communities and Equity Implications}

\titlerunning{Urban Boundaries, Social Barriers}

\author{Minwei Zhao\inst{1}\orcidlink{0000-0002-6380-5426}\thanks{These authors contributed equally.} \and
Weiming Zhang\inst{1}\orcidlink{0009-0003-2278}\textsuperscript{*} \and
Jiawang Du\inst{1}\orcidlink{0000-0002-1334-4158} \and
Qiming Liu\inst{2,1}\orcidlink{0009-0006-8421-0305} \and
Weiming Zhuang\inst{3}\orcidlink{0000-0001-8243-7772} \and
Pei Nie\inst{4}\orcidlink{0000-0003-2370-2112} \and
Cai Wu\inst{1}\orcidlink{0000-0002-5578-5525}\thanks{Corresponding author.}}

\authorrunning{M.~Zhao et al.}

\institute{The Hong Kong University of Science and Technology (Guangzhou)\\
\email{\{m.zhao,wzhang915,jdu146\}@connect.hkust-gz.edu.cn, caiwu@hkust-gz.edu.cn}
\and
School of Public Administration and Policy, Renmin University of China\\
\email{qimingliu937@ruc.edu.cn}
\and
Sony AI\\
\email{weiming.zhuang@sony.com}
\and
University of South China\\
\email{niepei@usc.edu.cn}}

\maketitle

\begin{figure}[!ht]
    \vspace{-10pt}
    \centering
    \includegraphics[width=1\linewidth]{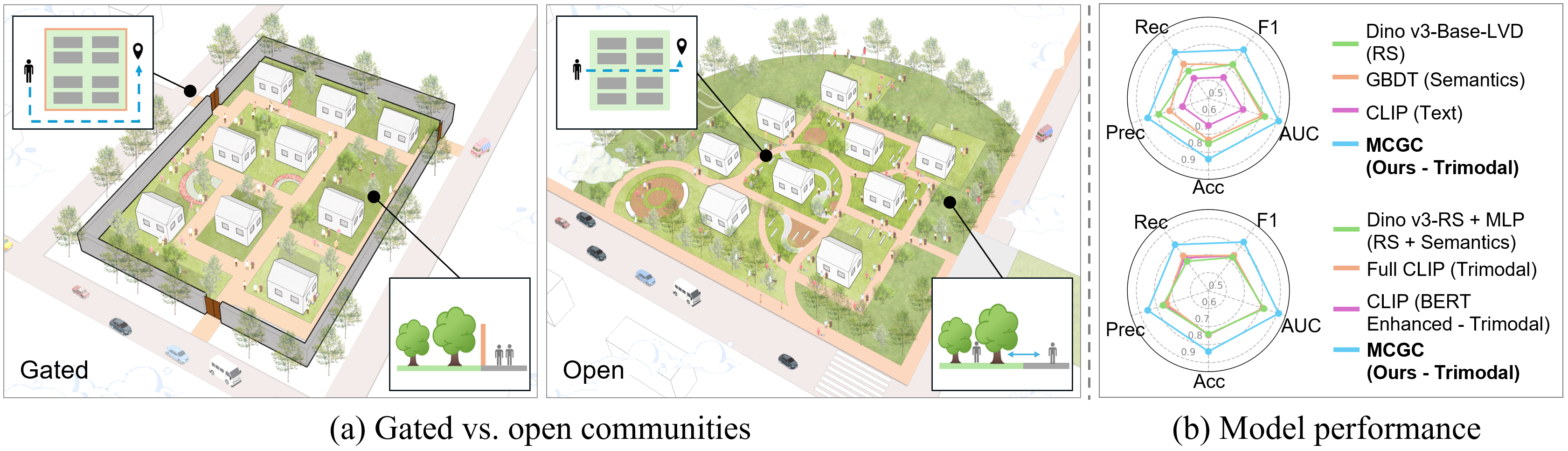}
    \small
    \caption{
    Comparison of gated vs. open communities and model performance. (a) Spatial form, route selection, and greenery accessibility differences between gated and open communities; (b) Superior community classification performance of \ours across five metrics compared to single-modality (top) and multimodal methods (bottom).}
    \vspace{-15pt}
    \label{fig:teaser_figure}
\end{figure}

\vspace{-15pt}
\begin{abstract}
Communities are fundamental spatial units that shape urban form and social life. Whether a residential compound is spatially open or enclosed affects mobility, access to public services, and equity, yet studies of Chinese \textit{fengbi xiaoqu} remain largely qualitative or small-scale, limiting reproducible city-scale analysis. We address this gap by introducing \textbf{GBA-GCs}, a metropolitan-scale multimodal benchmark for locally grounded gated/open community recognition in China’s Greater Bay Area, covering 37,444 residential compounds with aligned boundary polygons, high-resolution satellite imagery, Chinese metadata, and structured attributes, together with expert-verified labels, inter-annotator reliability, and official evaluation splits. Built on this benchmark, we present \textbf{Multimodal Classifier for Gated Community (MCGC)}, a vision-centric multimodal framework based on DINOv3-SAT that fuses imagery, text, and structured cues via modality-aware cross-attention and adaptive gating to mitigate modality imbalance. MCGC consistently outperforms strong unimodal and multimodal baselines. Finally, we apply the validated model to metropolitan-scale mapping and report equity-oriented findings including spatial clustering of GCs, privatized green space, and reduced pedestrian connectivity. The benchmark, code, and release documentation are available at \dataseturl.
  \keywords{Gated Community Detection\and Multimodal Learning\and Urban Computing\and Spatial Equity Analysis}
\end{abstract}

\section{Introduction}
\label{sec:intro}
\vspace{-1mm}
Recent advances in artificial intelligence and computer vision have profoundly transformed urban studies, enabling fine-grained mapping of urban morphology, land use, and environmental quality at unprecedented scales~\cite{wu2021convolutional,yuan2025survey}. These developments have bridged the gap between visual perception and spatial analytics, fostering a new paradigm of data-driven urban planning. However, many socially consequential urban forms remain underexplored from a visual computing perspective---most notably, Chinese \textbf{\textit{fengbi xiaoqu}} or gated residential compounds.

Gated communities mark a fundamental shift in urban aspirations toward security, exclusivity, and controlled access, reshaping both the spatial and social logics of contemporary cities~\cite{roitman2010gated,atkinson2005introduction,low2008gated}. In this paper, we do not impose a universal global taxonomy of ``gatedness''. Instead, we operationalize the Chinese \textit{fengbi xiaoqu}: residential compounds with continuous or near-continuous perimeters, limited or controlled entrances, and partial discontinuity from the public street network. This spatial form can alter patterns of accessibility, resource allocation, and social interaction~\cite{pow2015urban,breitung2012enclave}, often intensifying spatial segregation and exacerbating inequalities in mobility and public service provision~\cite{blakely1997divided,caldeira2000city,vesselinov2007gated} (Fig.~\ref{fig:teaser_figure}). 

Most existing studies on gated communities (GCs) rely on qualitative case analyses or small-scale mapping of tens to hundreds of neighborhoods~\cite{wu2004rise,qureshi2023gated,low2001edge}, limiting systematic generalization over large areas. This is largely due to fragmented data sources and the difficulty of distinguishing gated from non-gated communities at scale. Consequently, many computational urban analyses overlook or misclassify GCs, biasing estimates of walkability, green accessibility, and spatial equity~\cite{qing2011impact,webster2002global,tanulku2012gated,nicholls2001measuring}. For example, treating privatized GC greenery as public inflates perceived urban greenness, while ignoring GC boundaries underestimates road-network fragmentation~\cite{blinnikov2006gated,le2006gated}. These challenges motivate scalable, automated, and multimodal approaches that explicitly recognize gatedness as a spatial attribute.

To address this gap, we introduce \textbf{Multimodal Classifier for Gated Community (MCGC)} (Sec.~\ref{sec:MCGC}), which contributes both a new benchmark and a vision-centric multimodal architecture for Chinese enclosure recognition. On the data side, we compile the first \textbf{metropolitan-scale multimodal benchmark} (\textbf{GBA-GCs}) in China’s Greater Bay Area, covering \textbf{37,444} residential compounds with aligned boundaries, high-resolution satellite imagery, textual metadata, and structured attributes (e.g., FAR and inner POIs). We further standardize the task and evaluation protocol with expert-verified labels, inter-annotator reliability, and official splits, enabling fair and reproducible comparison for vision and multimodal methods in spatial computing. Dataset release, licensing, and access details are provided in Supplementary Sec. E.2.

On the methodological side, MCGC is a vision-centric multimodal framework tailored to the spatial logic of gatedness. Each community is represented by high-resolution imagery, semantic metadata, and structured attributes. We extract visual features with a DINOv3-SAT backbone (a satellite-adapted DINOv3~\cite{simeoni2025dinov3}), and encode text and numerical attributes with lightweight language and MLP modules. We align modalities via cross-modal attention and model interior--exterior context with a dual visual stream to capture fences, entrances, and boundary transitions. A modality-aware gating module further balances heterogeneous signals under real-world modality imbalance. Overall, MCGC consistently outperforms strong baselines, achieving about \textbf{+14.4\% F1} and \textbf{+12.2\% AUC} on 5-fold evaluation.

Beyond methodological performance, MCGC enables the first data-driven mapping of gated communities at a metropolitan-scale. The resulting benchmark reveals how enclosure patterns cluster spatially and interact with urban form, exposing the hidden social and environmental inequalities embedded in everyday urban structures.

In summary, our contributions are: \textbf{(I)} a large, metropolitan-scale multimodal benchmark for Chinese \textit{fengbi xiaoqu} recognition that combines high-resolution imagery, Chinese metadata, and structured attributes with expert-verified gated/open labels, inter-annotator reliability, official splits and evaluation protocol, forming a reproducible testbed;
\textbf{(II)} MCGC, a vision-centric multimodal framework (dual interior/exterior streams, modality-aware cross-attention, confidence-weighted fusion) fine-tuned from DINOv3-SAT, with consistent ablation gains and the strongest performance among unimodal and multimodal baselines;
\textbf{(III)} validated metropolitan-scale gated/open labels across the Greater Bay Area, yielding a region-wide enclosure map of 37{,}444 communities with human checks after inference; and
\textbf{(IV)} social good analyses revealing greenery bias (private greens), reduced pedestrian accessibility (barriers), and socioeconomic associations, calling for enclosure-aware metrics in planning, equity assessment, and spatial computing.
\vspace{-2mm}
\section{Related Work}
\vspace{-2mm}
\subsection{Gated Communities in Urban Studies}
Gated communities (GCs) have been extensively studied in urban sociology and planning, covering their historical emergence, social drivers, and implications for cohesion and governance~\cite{atkinson2005introduction,blakely1998separate,roitman2010gated,bagaeen2015beyond}. A consistent finding is that enclosure reinforces socio-spatial division and segregation~\cite{grant2004types,glasze2006segregation,he2013evolving,sharifi2013changes}, reduces cross-class interaction~\cite{coy2002gated,caldeira2000city,alkhafagie2024social}, and reshapes governance and resource allocation~\cite{bandauko2022systematic,chiu2023ending}. However, much of this literature relies on qualitative case studies or small-scale surveys spanning tens to hundreds of neighborhoods~\cite{wu2004rise,qureshi2023gated,low2001edge}.

In computational urban analytics, GCs are known to bias measurements of accessibility, green equity, and road-network connectivity~\cite{vesselinov2007gated,landry2009street,wu2020dismantling}: privatized greenery can inflate estimates of public environmental benefits~\cite{blinnikov2006gated,tanulku2012gated}, and ignoring enclosure boundaries can underestimate fragmentation and walkability constraints~\cite{qing2011impact,sun2020fine,switzky2001street}. Yet many computational studies still overlook or misclassify GCs as ordinary residential parcels, introducing systematic distortions in policy-relevant assessments of accessibility and spatial justice~\cite{blinnikov2006gated,landry2009street,wu2020dismantling,qing2011impact}. This gap persists partly because gatedness is difficult to represent as a scalable spatial attribute: existing attempts often rely on small curated samples or heuristic boundary identification~\cite{qing2011impact,le2008gated,xu2009gated}. As a result, large, accurately annotated, publicly available GC datasets, and scalable pipelines that incorporate enclosure morphology still remain limited.

\vspace{-5pt}
\subsection{Linking Gated Communities and Advances in AI and CV}
\vspace{-3pt}
Recent advances in AI, especially computer vision and multimodal learning, have reshaped urban analytics~\cite{bayoudh2022survey,ibrahim2021urban,deng2023survey}. High-resolution remote sensing enables scalable mapping of urban morphology~\cite{du2021mapping,sun2020fine}, and deep models achieve strong performance in land-use and functional zoning~\cite{cheng2018research,feng2018multi}. Self-supervised vision transformers (DINO/DINOv2) further improve transferability from unlabeled data~\cite{caron2021emerging,oquab2023dinov2,yin2024transformer,liu2025depth}, while foundation segmentation models (e.g., SAM) support flexible boundary delineation~\cite{kirillov2023segment,ma2024sam}.

However, many approaches remain vision-only and overlook the multimodal nature of urban environments, where imagery, textual metadata, and spatial structure interact~\cite{dalla2015challenges,huang2024crowdsourcing}. Although CLIP-style vision--language pretraining is powerful~\cite{radford2021learning}, its natural-image and caption pretraining limits performance on remote sensing and domain-specific semantics, making it unreliable for gatedness cues such as enclosure patterns, access control, and interior--exterior functional contrast. More broadly, gatedness is defined by discontinuities at community boundaries; capturing such transitions naturally calls for explicit interior--exterior modeling.

Motivated by these gaps, we present a unified benchmark-and-model pipeline for gated community recognition. Specifically, we release \textbf{GBA-GCs}, a metropolitan-scale multimodal benchmark that supports gated/open prediction under heterogeneous modality availability with expert-verified labels and fixed splits. Building on it, we develop \textbf{MCGC}, a vision-centric multimodal classifier that explicitly leverages interior--exterior visual context and integrates complementary non-visual cues, enabling reliable large-scale mapping for downstream urban equity analyses.
\begin{figure*}[!t]
    \centering
    \includegraphics[width=0.95\linewidth]{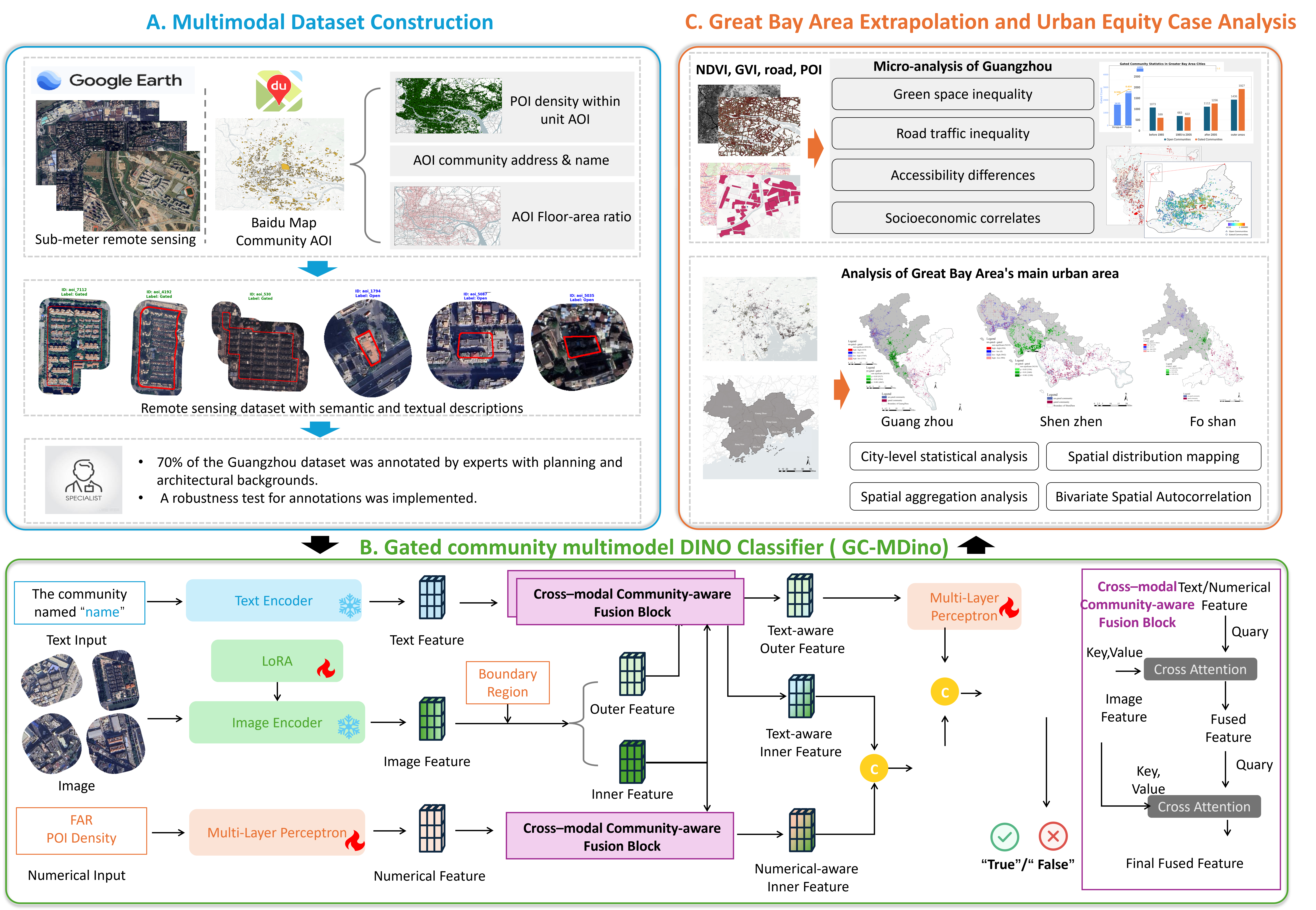}
    \vspace{-5pt}
    \small
    \caption{Overview of the MCGC framework.}
    \vspace{-16pt}
    \label{fig:pipeline}
\end{figure*}

\vspace{-2mm}
\section{Methodology}
\vspace{-2mm}
\subsection{Multimodal Benchmark Construction and Annotation Protocol}
\label{labelling}
As shown in Fig.~\ref{fig:pipeline}\textcolor{eccvblue}{A}, we construct \textbf{GBA-GCs}, a \textbf{metropolitan-scale multimodal benchmark} for recognizing Chinese \textit{fengbi xiaoqu} in GBA, consisting of 37,444 residential Areas of Interest (AOIs) with boundary polygons collected from licensed map-provider APIs~\cite{baidumap2025}. The AOIs cover the mapped residential communities in the region, enabling standardized and reproducible evaluation at metropolitan-scale. 
We formalize the benchmark task as binary classification (gated vs.\ open) under heterogeneous modality availability.
\paragraph{Standardized multimodal inputs.}
For each AOI, we retrieve 0.8\,m satellite imagery from Google Earth and crop tiles using a 50\,m contextual buffer around the polygon to preserve boundary evidence (e.g., walls, entrances, and edge transitions). In parallel, we query Baidu Maps for community metadata (name and address) and concatenate them into a short semantic text description. We additionally extract structured attributes (e.g., floor-area ratio and POI density) and standardize them into numerical features. 

\paragraph{Operational definition.}
We define a gated AOI as a residential compound with a continuous or near-continuous perimeter, limited or controlled entrances, and partial discontinuity between internal circulation and the surrounding public street network. We define an open AOI as a residential area that is publicly permeable, lacks compound-level access control, or is integrated with surrounding streets. This locally grounded definition is intended for Chinese \textit{fengbi xiaoqu}; external regions are used only as diagnostic transfer settings and not as evidence of a universal gatedness taxonomy.

\paragraph{Annotation protocol and quality control.}
To obtain high-quality ground truth for controlled evaluation, AOIs are annotated by domain experts with backgrounds in urban planning, architecture, and urban geography. Annotators assign gated/open labels by cross-checking satellite imagery, street-view evidence when available, community morphology, road connectivity, metadata, planning records, and textual descriptions under a standardized guideline. Ambiguous cases, including partial gating, mixed-use villages, occluded entrances, redevelopment sites, or conflicting source evidence, are escalated to senior adjudication under the same criteria. To quantify labeling reliability, we compute pairwise agreement and Cohen's $\kappa$ on a duplicated set of 200 AOIs:
\begin{subequations}
\begin{align}
\text{Agreement} &= \frac{1}{N}\sum_{i=1}^{N}\mathbf{1}\!\big(y_i^{(a)}=y_i^{(b)}\big), \\
\kappa &= \frac{p_o - p_e}{1 - p_e},
\end{align}
\end{subequations}
where $\mathbf{1}[\cdot]$ denotes the indicator function and $N$ is the number of duplicated samples; $p_o$ is the observed agreement and $p_e$ is the chance agreement estimated from annotators' marginal distributions.

\paragraph{Benchmark splits and evaluation protocol.}
To facilitate reproducible benchmarking, we provide official training/validation/test splits and a consistent evaluation protocol based on F1 and AUC (Sec.~\ref{sec:experiment}). Splits are constructed via joint stratification over AOI area and spatial distribution, ensuring coverage of diverse morphology (e.g., compound size and layout) and geographic contexts (e.g., inner city vs.\ suburban districts). These standardized inputs, expert labels, reliability checks, and fixed splits establish a unified testbed for gated-community recognition and downstream equity analyses.

\subsection{MCGC: A Multimodal Classifier for Gated Community}
\label{sec:MCGC}

To fully exploit the heterogeneous information contained in imagery, semantic text, and structured attributes, 
we propose MCGC, a multimodal architecture designed to capture enclosure-related cues through both 
interior--exterior contrast modeling and cross-modal community-aware fusion.  
As illustrated in Fig.~\ref{fig:pipeline}\textcolor{eccvblue}{B}, 
MCGC jointly encodes three modalities and learns their complementary interactions for robust gated community classification.

Given a textual input $\bm{x}_{\text{text}}$ (e.g., community name and designation), 
a satellite image $\bm{x}_{\text{img}}$, 
and structured numerical attributes $\bm{x}_{\text{num}}$ (e.g., FAR and POI density), 
we first extract modality-specific latent representations:
\begin{equation}
\begin{aligned}
\bm{F}_{\text{text}} &= \mathcal{E}_{\text{text}}(\bm{x}_{\text{text}}),\quad
\bm{F}_{\text{img}}  &= \mathcal{E}_{\text{img}}(\bm{x}_{\text{img}}),\quad
\bm{F}_{\text{num}}  &= \mathcal{E}_{\text{num}}(\bm{x}_{\text{num}}).
\end{aligned}
\end{equation}

Here, $\mathcal{E}_{\text{text}}$ is a frozen CLIP-style text encoder that captures semantic cues from community names; 
$\mathcal{E}_{\text{img}}$ is a pretrained DINOv3 encoder providing high-capacity visual representations; 
and $\mathcal{E}_{\text{num}}$ is a lightweight MLP mapping structured attributes into a compact feature space.  
To bridge the domain gap between natural-image DINOv3 features and our remote-sensing imagery, 
we apply LoRA adapters~\cite{hu2022lora} for efficient fine-tuning to better extract discriminative features from our remote-sensing imagery.

Since our task is a gated-community classification problem, the contrast between the community interior and its surrounding context is particularly informative: gated communities typically exhibit distinctive enclosure layouts and access-control-related visual patterns that manifest as appearance discrepancies across the boundary. 
Therefore, we further perform a boundary-aware decomposition of $F_{\text{img}}$ into an inner feature $F_{\text{inn}}$ and an outer feature $F_{\text{out}}$ by leveraging the annotated community boundary, which is also available as a structured attribute in $x_{\text{num}}$.
Specifically, we define a region-aware feature partition as:
\begin{equation}
F_{\text{inn}} = F_{\text{img}} \odot M, \qquad
F_{\text{out}} = F_{\text{img}} \odot (1-M),
\end{equation}
where $M\in\{0,1\}^{H\times W}$ is the binary mask of the community interior derived from the annotated boundary $B$, 
$1-M$ denotes the complementary exterior region, and $\odot$ denotes the Hadamard product applied channel-wise to the feature map.
We then compute a discrepancy feature:
\begin{equation}
F_{\text{disc}} = F_{\text{out}} - F_{\text{inn}}.
\end{equation}

Beyond cross-boundary contrast, MCGC further computes intra-region similarity statistics to capture lightweight structural cues that characterize the internal homogeneity and contextual consistency of different regions. For each region $m \in \{\text{img}, \text{inn}, \text{out}\}$,
\begin{equation}
S_{m} = 
\frac{1}{N_m^2} 
\sum_{i,j}
\frac{\bm{f}_{i}^{(m)} \cdot \bm{f}_{j}^{(m)}}{
\|\bm{f}_{i}^{(m)}\| \, \|\bm{f}_{j}^{(m)}\|},
\end{equation}
where $N_m$ denotes the number of feature vectors in region $m$; and 
$\bm{f}_i^{(m)}$ and $\bm{f}_j^{(m)}$ are the feature vectors at spatial locations $i$ and $j$. 
The expression computes the average pairwise cosine similarity within region $m$, producing a scalar measure of regional structural homogeneity. Thus, $S_{\text{img}}$, $S_{\text{inn}}$, and $S_{\text{out}}$ capture self-similarity in the global tile, the enclosed interior, and the surrounding exterior, respectively.
These scalar cues complement the learned representations by encoding how homogeneous or heterogeneous 
each region is, which is characteristic of enclosed residential forms.
\begin{figure}[t]
    \centering
    \vspace{-10pt}
    \includegraphics[width=0.6\linewidth]{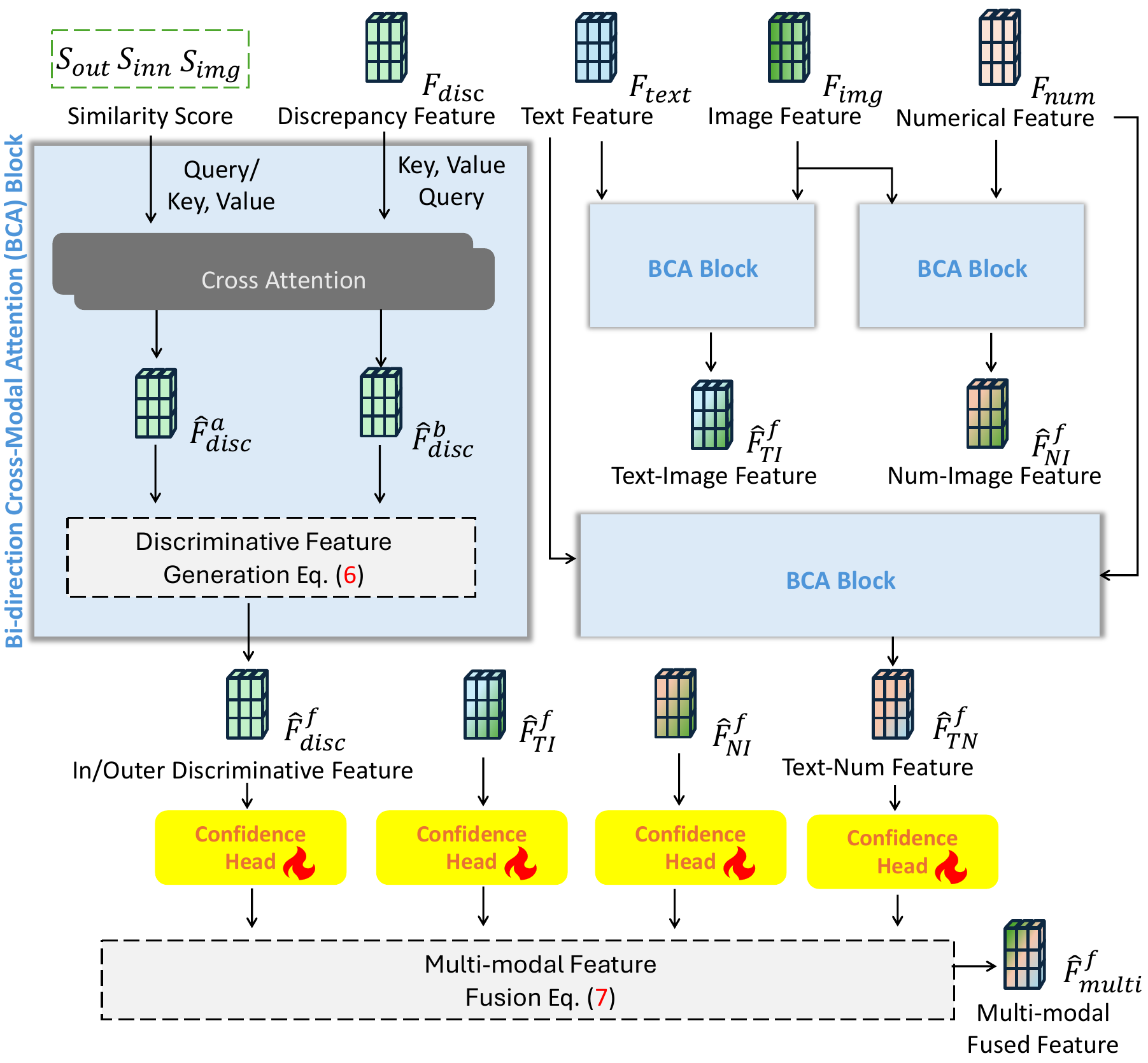}
    \caption{
        Overview of the Cross–modal Community-aware Fusion (CCF) Block.
    }
    \vspace{-15pt}
    \label{fig:bca_pipeline}
\end{figure}

\noindent\textbf{Cross-modal Community-aware Fusion (CCF) Block.}
The key component of MCGC is the CCF block, 
which performs cross-modal community-aware fusion by modeling the interactions among visual, textual, and numerical representations to produce a unified representation for gated-community recognition.  
CCF block is built upon a Bi-directional Cross-modal Attention (BCA) mechanism 
(Fig.~\ref{fig:bca_pipeline}), where each modality is alternately treated as query while the 
remaining modalities serve as key--value pairs. This reciprocal formulation enables the network to 
capture asymmetric yet complementary interactions across-modalities, a critical property when 
aligning imagery, text, and structured attributes with different semantic granularities.

To explicitly exploit enclosure-specific spatial cues, we jointly leverage the 
intra-region similarity statistics ($S_{\text{img}}, S_{\text{inn}}, S_{\text{out}}$) 
and the interior--exterior discrepancy feature $\bm{F}_{\text{disc}}$ within the BCA block. 
By coupling these two sources of information, the BCA block is able to emphasize 
boundary-aware patterns that are indicative of gated-community layouts.
Specifically, the BCA module performs bi-directional cross-attention between the 
structural cues and the discrepancy representation, producing two directional 
attentional embeddings, $\hat{\bm{F}}^{a}_{\text{disc}}$ and 
$\hat{\bm{F}}^{b}_{\text{disc}}$, corresponding to the two query directions. 
These embeddings encode complementary cross-modal correlations between regional 
structure statistics and boundary discrepancy signals.
To consolidate these directional responses, we introduce a soft alignment mechanism 
that measures their mutual consistency and reweights the features accordingly:
\begin{equation}
\hat{\bm{F}}^{f}_{\text{disc}}
=
\operatorname{softmax}\!\left(
\frac{
\hat{\bm{F}}^{a}_{\text{disc}}
\hat{\bm{F}}^{b^{\!\top}}_{\text{disc}}
}{
\sqrt{d}
}
\right)
\hat{\bm{F}}^{b}_{\text{disc}},
\end{equation}
where $\hat{\bm{F}}^{a}_{\text{disc}}$ and $\hat{\bm{F}}^{b}_{\text{disc}} 
\in \mathbb{R}^{N \times d}$ denote the two directional attention embeddings. 
The matrix product $\hat{\bm{F}}^{a}_{\text{disc}}\hat{\bm{F}}^{b^{\!\top}}_{\text{disc}}$ 
measures pairwise affinity between the two directional representations, allowing 
the model to highlight mutually reinforcing responses while attenuating inconsistent 
cross-modal signals.
The resulting feature $\hat{\bm{F}}^{f}_{\text{disc}}$ therefore serves as a 
dedicated \emph{interior--exterior discriminative representation}, which encodes 
boundary-aware enclosure contrast in a cross-modally aligned feature space.

Beyond the specialized difference pathway, the CCF block also processes three modality pairs  
(text--image, numerical--image, and text--numerical) through independent BCA modules, producing  
$\hat{\bm{F}}^{f}_{\text{TI}}$, $\hat{\bm{F}}^{f}_{\text{NI}}$, and $\hat{\bm{F}}^{f}_{\text{TN}}$.  
Each fused pairwise representation is passed through a lightweight confidence head that predicts a 
learnable modality weight:
\begin{equation}
\hat{\bm{F}}^{f}_{\text{multi}}
= 
\sum_{m} 
\alpha_m \hat{\bm{F}}^{f}_{\text{m}}
\quad
\alpha_m = \operatorname{softmax}(h_{\text{conf}}( \hat{\bm{F}}^{f}_{\text{m}})),
\end{equation}
allowing the model to adaptively modulate the contribution of each modality based on its reliability.  
The final fused feature $\hat{\bm{F}}^{f}_{\text{multi}}$ is fed into a classification head to predict whether 
a community is gated or open.

\subsection{Gated Community in the Greater Bay Area}
To investigate enclosure patterns at a metropolitan-scale, we apply the trained MCGC classifier to the full AOI dataset across the Greater Bay Area (GBA), covering nine mainland cities. The dataset contains \textbf{37,444 residential communities}, representing the residential fabric of the region. Model inference produces region-wide gatedness labels for every AOI, resulting in the first open-access, high-resolution map of residential enclosures across the mainland GBA.

To ensure the robustness of these large-scale outputs, we adopt the same verification protocol used in benchmark construction (Sec.~\ref{labelling}). A post-inference consistency assessment compares the automatically predicted labels with independent human verification, providing a direct measure of reliability and identifying potential boundary cases for refinement.

With the validated dataset in place, we conduct several proof-of-concept analyses to illustrate its analytical value. At the inter-city level, we quantify cross-city differences in enclosure prevalence, using indicators such as the number, share, and spatial footprint of gated communities. At the intra-city level, we examine how gating correlates with urban development, greenery, circulation structure,  and socioeconomic segregation. These explorations demonstrate how metropolitan-scale enclosure data can support comparative urban morphology research and deepen understanding of enclosure dynamics. Detailed analytical procedures are provided in Supplementary Sec. B, data licensing and reproducibility details are provided in Supplementary Sec. E.



\begin{table*}[!t]
\centering
\vspace{-10pt}
\caption{ Performance comparison between MCGC and a comprehensive set of unimodal and multimodal baselines on gated community classification. Since no prior work addresses this task, all baselines are implemented using standard backbones (CNN, CLIP, DINOv3, BERT, MLP) and straightforward fusion strategies (early concatenation or linear fusion); detailed model descriptions are in Supplementary Sec. D. \textbf{MCGC achieves the best overall performance across Acc, Prec, F1, and AUC}, and substantially outperforms both unimodal and multimodal fusion baselines. Although the MLP model attains the highest recall due to strong bias toward the positive class, its precision and overall F1 are significantly lower than MCGC. }
\resizebox{0.95\textwidth}{!}{
\begin{tabular}{lcccccc}
\toprule
\textbf{Model} & \textbf{Acc} & \textbf{Prec} & \textbf{Rec} & \textbf{F1} & \textbf{AUC} & \textbf{Modalities} \\
\midrule
SimpleCNN-N2N & 0.731 \tiny{± 0.005} & 0.730 \tiny{± 0.034} & 0.675 \tiny{± 0.061} & 0.698 \tiny{± 0.020} & 0.805 \tiny{± 0.007} & Remote Sensing (RS) \\
SimpleCNN-Frozen & 0.644 \tiny{± 0.018} & 0.691 \tiny{± 0.038} & 0.519 \tiny{± 0.084} & 0.586 \tiny{± 0.049} & 0.718 \tiny{± 0.020} & RS \\
CLIP ViT-B/16 & 0.752 \tiny{± 0.015} & 0.769 \tiny{± 0.025} & 0.716 \tiny{± 0.049} & 0.740 \tiny{± 0.023} & \underline{0.838 \tiny{± 0.014}} & RS \\
DINO v2-Small & 0.740 \tiny{± 0.003} & 0.739 \tiny{± 0.034} & 0.742 \tiny{± 0.071} & 0.737 \tiny{± 0.019} & 0.826 \tiny{± 0.011} & RS \\
DINO v3-Base-LVD & \underline{0.753 \tiny{± 0.010}} & 0.789 \tiny{± 0.026} & 0.687 \tiny{± 0.052} & 0.732 \tiny{± 0.021} & 0.830 \tiny{±  0.012} & RS \\
DINOv3-SAT-Fixed & 0.747 \tiny{± 0.018} & \underline{0.800 \tiny{± 0.036}} & 0.657 \tiny{± 0.066} & 0.718 \tiny{± 0.031} & 0.833 \tiny{±  0.011} & RS \\
MLP & 0.571 \tiny{± 0.054} & 0.540 \tiny{± 0.035} & \textbf{0.966} \tiny{± 0.030} & 0.691 \tiny{± 0.021} & 0.790 \tiny{± 0.018} & Numerical \\
GBDT & 0.731 \tiny{± 0.015} & 0.726 \tiny{± 0.014} & 0.734 \tiny{± 0.020} & 0.730 \tiny{± 0.016} & 0.813 \tiny{± 0.016} & Numerical \\
BERT Chinese & 0.695 \tiny{± 0.019} & 0.704 \tiny{± 0.012} & 0.663 \tiny{± 0.029} & 0.682 \tiny{± 0.020} & 0.759 \tiny{± 0.015} & Text \\
CLIP Text OpenAI & 0.651 \tiny{± 0.017} & 0.651 \tiny{± 0.018} & 0.637 \tiny{± 0.032} & 0.643 \tiny{± 0.021} & 0.702 \tiny{± 0.023} & Text \\
DINOv3-SAT+ BERT Chinese & 0.740 \tiny{± 0.014} & 0.743 \tiny{± 0.016} & 0.729 \tiny{± 0.062} & 0.739 \tiny{± 0.028} & 0.823 \tiny{± 0.013}  & RS + Text\\
DINOv3-SAT + MLP & 0.742 \tiny{± 0.012} & 0.765 \tiny{± 0.047} & 0.701 \tiny{± 0.056} & 0.728 \tiny{± 0.016} & 0.824 \tiny{± 0.013}  & RS + Numerical\\
BERT Chinese + MLP & 0.717 \tiny{± 0.027} & 0.692 \tiny{± 0.047} & 0.788 \tiny{± 0.059} & 0.734 \tiny{± 0.015} & 0.807 \tiny{± 0.022}  & Text + Numerical \\
Full CLIP (Three-modal) & 0.744 \tiny{± 0.013} & 0.743 \tiny{± 0.018} & 0.738 \tiny{± 0.005} & \underline{0.741 \tiny{± 0.010}} & 0.819 \tiny{± 0.017}  & RS + Text + Numerical \\
CLIP (BERT Enhanced) & 0.746 \tiny{± 0.013} & 0.753 \tiny{± 0.014} & 0.724 \tiny{± 0.014} & 0.739 \tiny{± 0.013} & 0.818 \tiny{± 0.014}  & RS + Text + Numerical \\
\textbf{MCGC (Ours)} &\textbf{0.853 \tiny{±0.037}} & \textbf{0.869 \tiny{±0.052}} & \underline{0.830} \tiny{±0.033} & \textbf{0.848 \tiny{±0.037}} & \textbf{0.917 \tiny{±0.039}} & RS + Text + Numerical \\
\midrule
\textbf{Improvement} & \textbf{+14.6}\% & \textbf{+17.0}\% & \textbf{+12.4}\% & \textbf{+14.4}\% & \textbf{+12.2}\% & Over CLIP (Three-modal) \\
\bottomrule

\end{tabular}
}
\vspace{-10pt}
\label{tab:baseline_gc}
\end{table*}

\begin{table*}[!t]
\centering
\caption{Ablation study of key components in MCGC. Metrics are averaged over five stratified 8:2 splits with mean ± standard deviation. 
$\Delta$ values are computed relative to the previous row. 
$\Delta \overline{\boldsymbol{\sigma}}$ denotes the change in mean standard deviation across Acc/Prec/Rec/F1/AUC (positive = less stable, negative = more stable).}
\label{tab:ablation_gc_deltasigma}
\setlength{\tabcolsep}{4pt}
\resizebox{0.95\textwidth}{!}{
\begin{tabular}{lcccccccc}
\toprule
\textbf{Model Variant} & \textbf{Acc} & \textbf{Prec} & \textbf{Rec} & \textbf{F1} & \textbf{AUC} & $\mathbf{\Delta}$\textbf{F1} & $\mathbf{\Delta}$\textbf{AUC} & $\Delta \overline{\boldsymbol{\sigma}}$ \\
\midrule
DINOv3-SAT (Frozen) & 0.747 \tiny{±0.018} & 0.800 \tiny{±0.036} & 0.657 \tiny{±0.066} & 0.718 \tiny{±0.031} & 0.833 \tiny{±0.011} & -- & -- & -- \\
+ LoRA Fine-tuning & 0.779 \tiny{±0.012} & 0.801 \tiny{±0.035} & 0.739 \tiny{±0.034} & 0.768 \tiny{±0.011} & 0.850 \tiny{±0.016} & $+0.050$ & $+0.017$ & $-0.004$ \\
+ BERT-Chinese (Text) & 0.823 \tiny{±0.040} & 0.845 \tiny{±0.038} & 0.786 \tiny{±0.067} & 0.813 \tiny{±0.046} & 0.902 \tiny{±0.039} & $+0.045$ & $+0.052$ & $+0.024$ \\
+ Numerical Attributes & 0.824 \tiny{±0.026} & 0.840 \tiny{±0.044} & 0.799 \tiny{±0.026} & 0.818 \tiny{±0.024} & 0.902 \tiny{±0.023} & $+0.005$ & $+0.000$ & $-0.017$ \\
+ Three-modal CCF Block (Fusion) & \underline{0.838} \tiny{±0.031} & \underline{0.854} \tiny{±0.047} & \underline{0.815 \tiny{±0.050}} & \underline{0.833} \tiny{±0.033} & \underline{0.910} \tiny{±0.028} & $+0.015$ & $+0.008$ & $+0.009$ \\
+ Dual Visual Stream (IO) & \textbf{0.853 \tiny{±0.037}} & \textbf{0.869 \tiny{±0.052}} & \textbf{0.830} \tiny{±0.033} & \textbf{0.848 \tiny{±0.037}} & \textbf{0.917 \tiny{±0.039}} & $+0.015$ & $+0.015$ & $-0.005$ \\
\bottomrule
\end{tabular}
} 
\vspace{-12pt}
\end{table*}

\section{Experiments}
\label{sec:experiment}
\vspace{-1mm}
\subsection{Experimental Setup}
\noindent\textbf{Dataset (Benchmark and evaluation protocol).}
We evaluate MCGC under the standardized setting of GBA-GCs, a metropolitan-scale multimodal benchmark covering 37,444 residential AOIs across China’s Greater Bay Area. 
To enable reproducible and controlled benchmarking, we report quantitative results on an expert-annotated evaluation split from Guangzhou, where each AOI is labeled as gated or open following the protocol in Sec.~\ref{labelling}. 
Train, validation and test splits are constructed via joint stratification by AOI area and spatial distribution, ensuring balanced coverage of diverse urban morphologies and geographic contexts across splits. 
Labeling consistency is assessed on a duplicated set re-labeled by all annotators, achieving 94\% agreement with Cohen’s $\kappa{=}0.85$. 
The resulting Guangzhou evaluation split is class-balanced, containing 2{,}663 open (50.6\%) and 2{,}605 gated (49.4\%) communities. A benchmark card summarizing dataset scope, modalities, and evaluation protocol is provided in Supplementary Sec. E.1, and the fixed stratified splits and random seeds are detailed in Supplementary Sec. D.4.

\noindent\textbf{GBA Gated community map.}
Beyond benchmark evaluation, we apply the trained model to all AOIs in the full GBA-GCs benchmark to produce a region-wide gated/open map (Sec.~\ref{sec:Analysis}). 
To verify robustness of the inferred labels outside Guangzhou, we conduct a post-hoc human check on 200 AOIs sampled from Shenzhen and Foshan, where inferred labels reach 88\% agreement with expert verification ($\kappa{=}0.75$). Similarly, the agreement in Hong Kong subset reaches 95\% (Cohen’s $\kappa$ = 0.73), indicating the generalizability of the data across diverse spatial development contexts. 
These checks support the reliability of the metropolitan-scale mapping results, which we treat as an additional contribution distinct from the controlled benchmark evaluation. We further provide a failure and borderline case library in Supplementary Sec. D.8, covering common ambiguity types and representative false-positive/false-negative patterns.

\noindent\textbf{Implementation Details and Evaluation Metrics.} Implementation details and evaluation protocol are provided in Supplementary Sec. D.5. We adopt standard binary classification metrics, including Accuracy (Acc), Precision (Prec), Recall (Rec), F1-score (F1), and the Area Under the ROC Curve (AUC). 

\vspace{-2mm}
\subsection{Baseline Comparison and Discussion}
\vspace{-1mm}
Tab.~\ref{tab:baseline_gc} compares MCGC with unimodal and multimodal baselines for gated community detection. Vision-only backbones (DINOv2, DINOv3-SAT) achieve strong performance (F1 $\approx$ 0.72--0.74), indicating that remote-sensing representations capture enclosure cues such as boundaries and edge transitions. In contrast, text-only encoders (BERT-Chinese, CLIP-text) underperform, consistent with sparse and noisy community metadata. Models using only structured attributes show limited standalone predictive power.

Na\"ive multimodal fusion (early concatenation) provides marginal or inconsistent gains over vision-only models, while CLIP-based variants yield moderate AUC improvements but remain limited on high-resolution satellite textures. Overall, these results suggest that effective gatedness recognition benefits from \emph{boundary-conditioned visual modeling} and \emph{careful multimodal interaction} rather than simple feature aggregation. We do not interpret the gain as evidence that exact polygon shape is essential; instead, the task requires a meaningful AOI footprint and interior--exterior context, as tested further in Sec.~\ref{sec:robustness}.

\begin{figure}[t]
  \centering
  \begin{subfigure}[t]{0.42\linewidth}
    \centering
    \includegraphics[width=\linewidth]{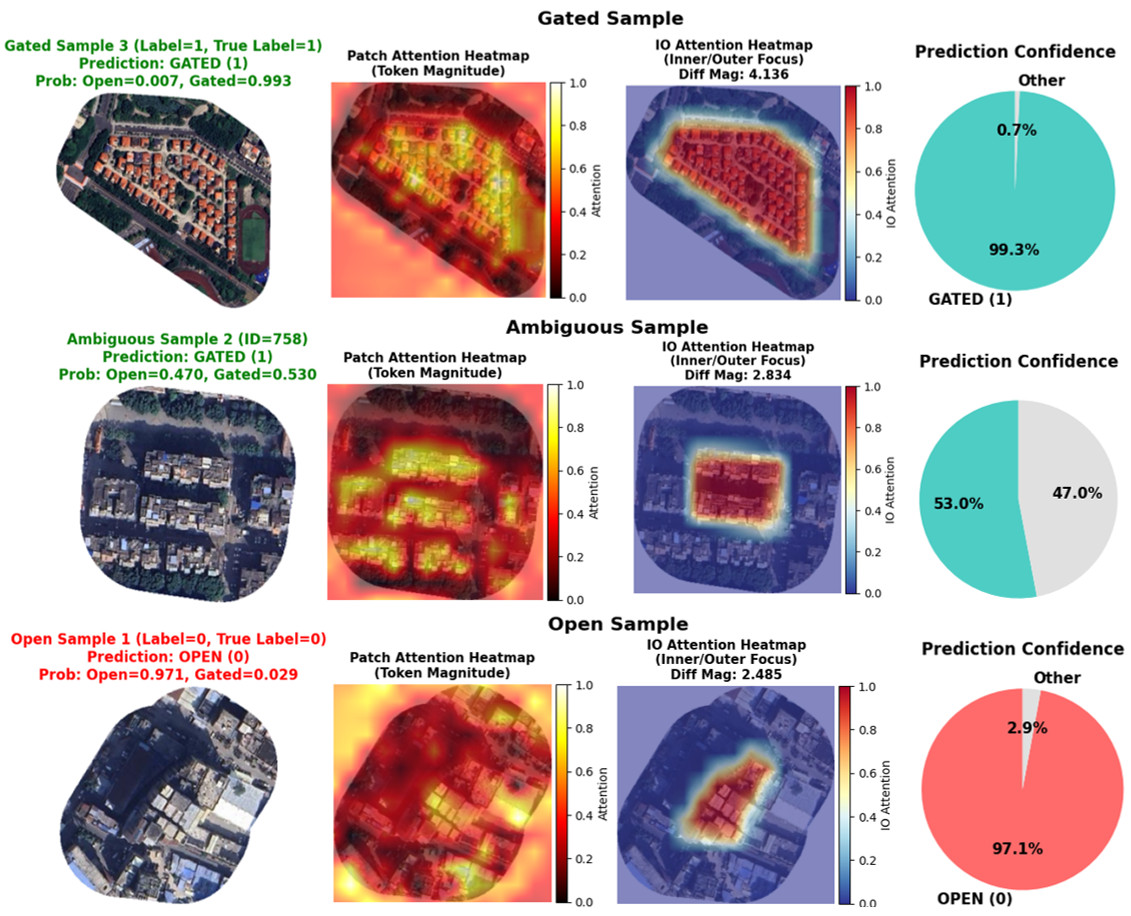}
    \caption{\scriptsize Representative inference cases of MCGC.}
    \label{fig:case_study_main}
  \end{subfigure}
  \hfill
  \begin{subfigure}[t]{0.55\linewidth}
    \centering
    \includegraphics[width=\linewidth]{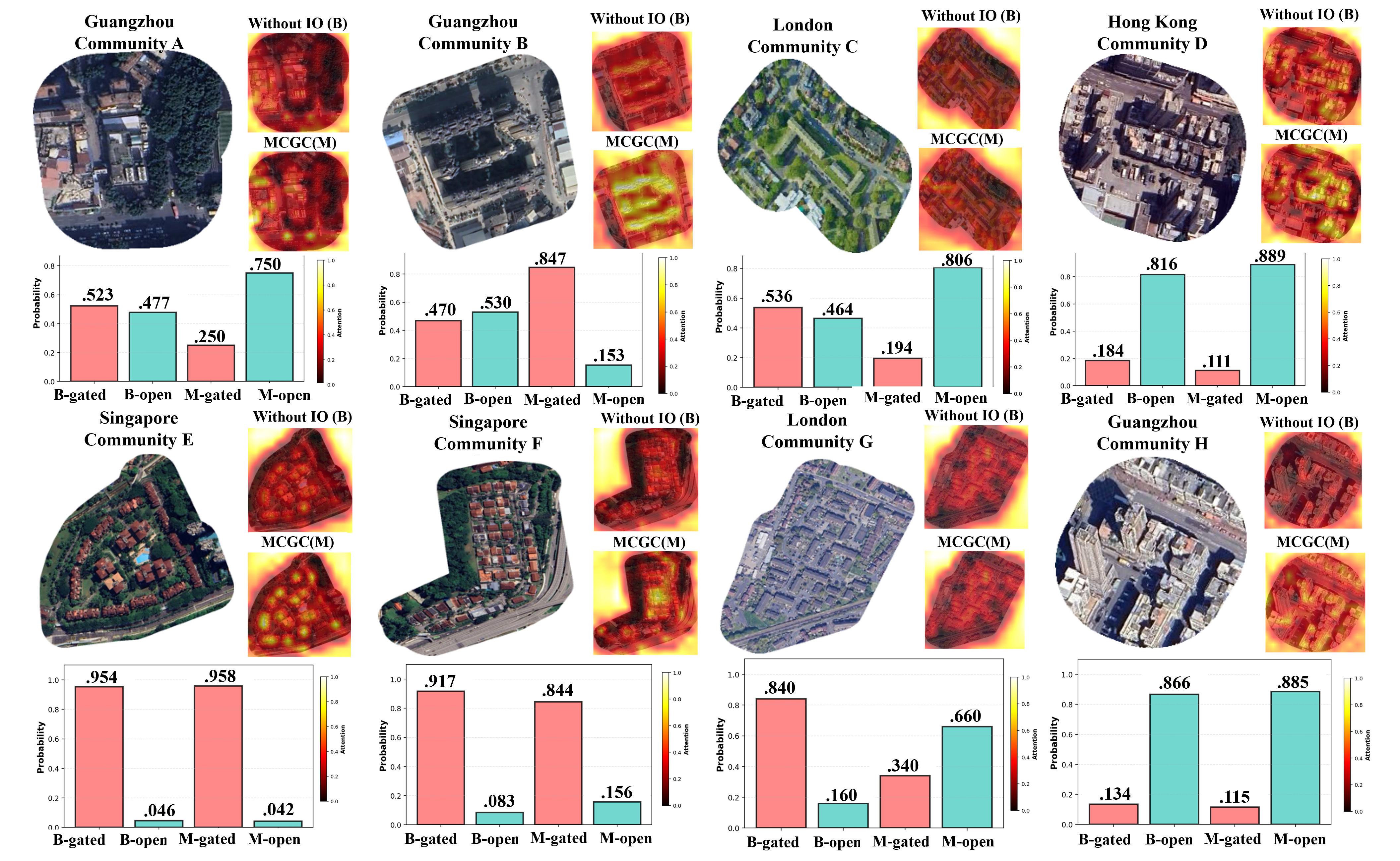}
    \caption{\scriptsize Case studies. MCGC corrects predictions via boundary-aware visual cues, zoom in for better review.}
    \label{fig:misleading_meta}
  \end{subfigure}
  \vspace{-2mm}
  \caption{\textbf{Qualitative evidence for boundary-aware interpretability.}
  Left: representative cases.
  Right: failure modes under misleading metadata, where IO-aware visual cues help correct predictions (probabilities shown).}
  \label{fig:qualitative_io_cases}
  \vspace{-20pt}
\end{figure}

To isolate the contribution of each component, we conduct ablations on the annotated Guangzhou benchmark (Tab.~\ref{tab:ablation_gc_deltasigma}). LoRA fine-tuning consistently improves the frozen DINOv3-SAT backbone, with gains largely driven by higher recall, suggesting better adaptation to local urban morphology. Adding BERT-Chinese metadata yields the most pronounced improvement in separability (AUC), while also increasing variance across splits, consistent with sparse and noisy text availability. Numerical attributes provide only marginal gains in mean performance but noticeably reduce run-to-run variance, indicating a stabilizing effect from structured cues. Building on these inputs, the three-modal CCF module further boosts both F1 and AUC by learning modality-aware interactions beyond early concatenation. Finally, the dual visual stream (IO) improves overall accuracy and F1 while reducing variance, supporting its role in robustness for boundary-ambiguous cases. We further provide qualitative evidence in Fig.~\ref{fig:qualitative_io_cases}. Attention maps highlight boundary, entrance and building regions, and cross-region cases illustrate that the IO stream can recover borderline predictions by leveraging interior--exterior contrast.

\begin{table}[t]
\centering
\small
\setlength{\tabcolsep}{5pt}
\renewcommand{\arraystretch}{1.02}

\caption{\textbf{Robustness analyses.}
\textbf{(a)} Text reliance ablation under controlled perturbations of textual metadata.
\textbf{(b)} Cross-region generalization for direct transfer from Guangzhou to external regions under heterogeneous AOI
quality (\textbf{AOI}: exact boundaries (Y) vs.\ inexact public footprints (N)).
$B^{*}$: na\"ive multimodal fusion baseline; $IO^{-}$: MCGC without the IO stream; $IO^{\wedge}$: full MCGC with IO
(``w/o IO''/``+ IO'' in Sec.~4.2 correspond to $IO^{-}$/$IO^{\wedge}$).
\textbf{(c)} Boundary perturbation robustness under AOI mask translation/erosion/dilation (erosion/dilation radius scaled
by $\sqrt{\text{area}}$).}
\label{tab:robustness_all}
\vspace{-4pt}

\begin{minipage}[t]{0.4\linewidth}
\centering
\setlength{\tabcolsep}{4pt}
\resizebox{0.9\linewidth}{!}{
\begin{tabular}{lccc}
\toprule
\textbf{Setting} & \textbf{ACC} & \textbf{F1} & \textbf{AUC} \\
\midrule
Text-only (BERT-C) & 0.695 & 0.682 & 0.759 \\
Text-only (CLIP)   & 0.695 & 0.682 & 0.702 \\
MCGC (BERT-C)      & \textbf{0.853} & \textbf{0.848} & \textbf{0.917} \\
MCGC (CLIP)        & 0.818 & 0.814 & 0.882 \\
w/o Text (Test)    & 0.810 & 0.797 & 0.886 \\
Shuffle Text (Test)& 0.809 & 0.797 & 0.886 \\
w/o Text (T+T)     & 0.791 & 0.787 & 0.891 \\
Shuffle Text (T+T) & 0.777 & 0.763 & 0.857 \\
\bottomrule
\end{tabular}
}
\vspace{2pt}
{\footnotesize \textbf{(a)} Text reliance ablation.}
\end{minipage}
\hfill
\begin{minipage}[t]{0.48\linewidth}
\centering
\setlength{\tabcolsep}{4pt}
\resizebox{1\linewidth}{!}{
\begin{tabular}{l c l c c c}
\toprule
\textbf{Region} & \textbf{AOI} & \textbf{Metric} & $B^{*}$ & $IO^{-}$ & $IO^{\wedge}$ \\
\midrule
Hong Kong  & Y & Acc & 0.666 & 0.745 & \textbf{0.863} \\
Hong Kong  & Y & AUC & 0.835 & 0.909 & \textbf{0.925} \\
\midrule
Singapore  & N & Acc & 0.796 & 0.843 & \textbf{0.875} \\
Singapore  & N & AUC & \textbf{0.745} & 0.702 & 0.732 \\
\midrule
London     & N & Acc & 0.558 & \textbf{0.674} & \textbf{0.674} \\
London     & N & AUC & 0.328 & 0.595 & \textbf{0.627} \\
\bottomrule
\end{tabular}}
\vspace{5pt}
{\footnotesize \textbf{(b)} Cross-region generalization.}
\end{minipage}

\begin{minipage}[t]{0.88\linewidth}
\centering
\setlength{\tabcolsep}{5pt}
\resizebox{1\linewidth}{!}{
\begin{tabular}{lcccc}
\toprule
\textbf{Perturbation} & \textbf{1\,m / 1\%} & \textbf{2\,m / 5\%} & \textbf{5\,m / 10\%} & \textbf{10\,m / 20\%} \\
\midrule
Translation (m) & 0.808 / 0.798 & 0.798 / 0.789 & 0.772 / 0.765 & 0.738 / 0.732 \\
Erosion (\%)    & 0.811 / 0.801 & 0.798 / 0.788 & 0.778 / 0.771 & 0.752 / 0.745 \\
Dilation (\%)   & 0.809 / 0.797 & 0.792 / 0.783 & 0.768 / 0.759 & 0.741 / 0.734 \\
\bottomrule
\end{tabular}
}
\vspace{2pt}
{\footnotesize \textbf{(c)} Boundary perturbation robustness on Guangzhou (ACC/F1).}
\end{minipage}
\vspace{-15pt}
\end{table}

\vspace{-2mm}
\subsection{Robustness Analyses}
\vspace{-1mm}
\label{sec:robustness}
\noindent\textbf{Robustness to modality noise.}
As shown in Tab.~\ref{tab:ablation_gc_deltasigma}, introducing BERT-Chinese metadata yields a notable performance gain but also increases variance across splits, raising the question of whether the model becomes overly sensitive to noisy textual cues. We therefore conduct controlled text perturbations to assess potential shortcut reliance. As shown in Tab.~\ref{tab:robustness_all}\textbf{(a)}, removing or shuffling text \emph{only at test time} causes minor changes, suggesting that MCGC does not depend on metadata as a dominant cue. In contrast, removing/shuffling text during \emph{both} training and testing leads to a larger drop, indicating that text primarily serves as a complementary signal that improves learning when available.

\noindent\textbf{Cross-region generalization under heterogeneous AOI quality.}
We further evaluate direct transfer from Guangzhou to three external regions with expert-verified labels:
\textbf{Hong Kong} (AOI=Y), \textbf{Singapore} (AOI=N), and \textbf{London} (AOI=N), without any target-domain fine-tuning.
Tab.~\ref{tab:robustness_all}\textbf{(b)} shows strong transfer performance, especially when accurate AOIs are available
(Hong Kong), and competitive results under inexact footprints (London). On Singapore, the full model improves accuracy
while the na\"ive baseline attains a slightly higher AUC; this is expected since AUC can be more sensitive to ranking and
calibration under noisy delineations, whereas boundary-aware cues primarily benefit thresholded decisions used for
region-wide mapping.

\noindent\textbf{Robustness to boundary noise.}
Finally, we perturb AOI masks at test time via translation/erosion/dilation while keeping the image tile fixed.
Tab.~\ref{tab:robustness_all}\textbf{(c)} shows that performance degrades gracefully as boundary noise increases, and even under the strongest perturbations
(10\,m translation or 20\% erosion/dilation), the drop is limited.
This suggests that the IO stream provides a reliable structural cue and that MCGC does not require extremely precise AOI delineation to function effectively, which is important for scaling to diverse regions with heterogeneous mapping quality.
Additional geometry controls further separate boundary location from exact shape. When the AOI location is corrupted, F1 drops from .849 to .797/.801, and removing exterior context also reduces performance (.805). In contrast, random-shape and bounding-box masks at the correct approximate location remain nearly tied with the true mask (.848 vs.\ .849). We therefore frame IO as a useful boundary-conditioned inductive bias: approximate AOI location and interior--exterior contrast matter, while precise polygon shape alone is not the decisive signal.

\vspace{-2mm}
\subsection{Extended Dataset and Exploratory Analysis}
\label{sec:Analysis}
\vspace{-1mm}
Building on the validated MCGC framework, we extend model inference to the Greater Bay Area (GBA), covering nine mainland cities. This process produces the first region-wide dataset of gated and open residential communities across mainland GBA, released for open research use. Leveraging this dataset, we conduct four exploratory analyses---temporal--spatial patterns, pedestrian accessibility impacts, greenery perception bias, and socioeconomic associations---to illustrate how fine-grained enclosure data can support broader social-good research on segregation, spatial governance, social equity, and spatial justice. A summary of the dataset, along with detailed visualizations and statistical reports for the exploratory analyses, is provided in Supplementary Secs. A and C.
\noindent\textbf{Temporal–Spatial Patterns of Gated Communities.}
\begin{figure}[t]
  \centering
  \begin{subfigure}[t]{0.55\linewidth}
    \centering
    \includegraphics[width=\linewidth]{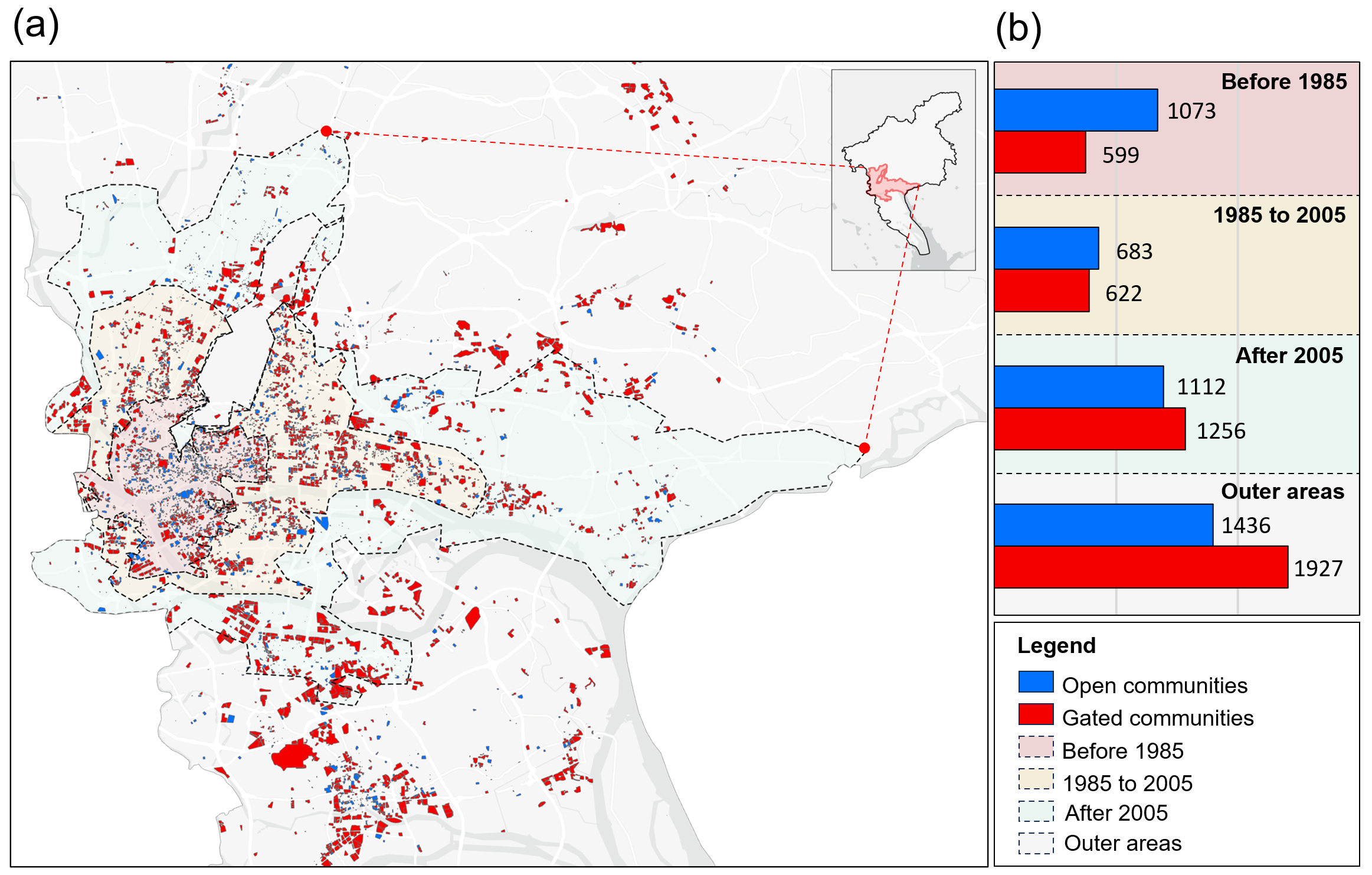}
    \caption{\scriptsize cohort-based temporal--spatial analysis obtained by intersecting present-day gated/open labels with historical development cohorts.}
    \label{fig:ts_patterns}
  \end{subfigure}
  \hfill
  \begin{subfigure}[t]{0.41\linewidth}
    \centering
    \includegraphics[width=\linewidth]{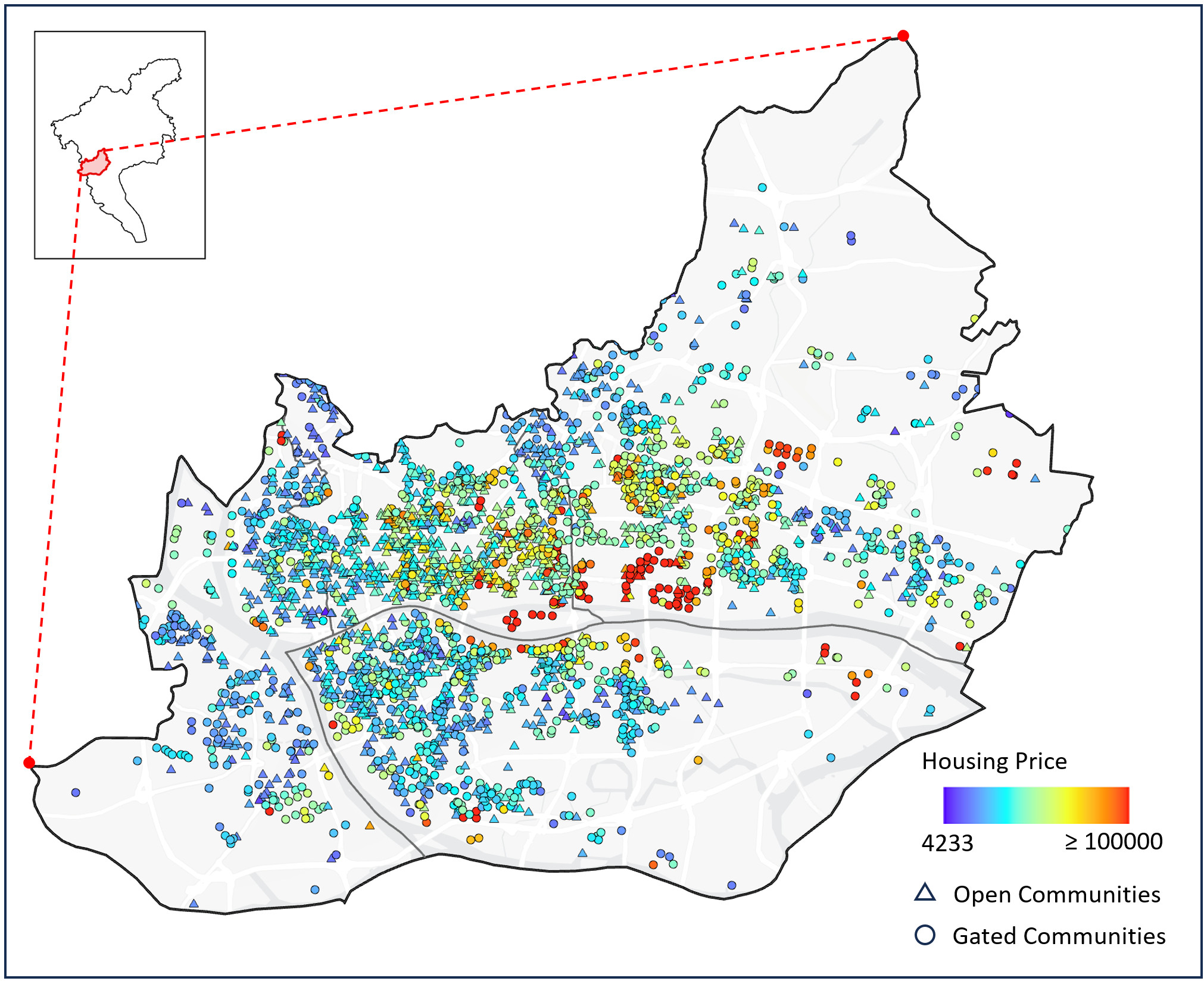}
    \caption{\scriptsize housing-price stratification of gated vs.\ open communities in central districts.}
    \label{fig:price_patterns}
  \end{subfigure}
  \vspace{-2mm}
  \caption{\textbf{Exploratory analyses enabled by region-wide enclosure mapping.}}
  \label{fig:expeditionary_maps}
  \vspace{-3mm}
\end{figure}
Using Guangzhou as a representative case, we observe a pronounced shift from open to gated residential forms (Fig.~\ref{fig:expeditionary_maps}\textbf{(left)}). Open communities dominated before 1985, but after 2005 gated enclaves became prevalent, especially in rapidly expanding peripheral districts. This transition aligns with the rise of large superblocks and low-permeability layouts, reflecting a broader move toward more enclosed and spatially segregated neighborhood structures.

\noindent\textbf{Pedestrian Accessibility and Detour Effects.}
When gated perimeters are accounted for, simulated routes to major public service POIs exhibit significant detour penalties (Tab.~\ref{tab:equity_tables}\textbf{(c)}). Around Taikoo Hui, the average Detour Index (DI) reaches 1.164 (max 1.477), adding more than 225\,m of travel. Around the Tianhe Women and Children’s Hospital, DI averages 1.102 (max 1.239) with nearly 150\,m in extra distance. Across both sites, over 70\% of origins have DI \textgreater{} 1.05, revealing widespread distortion of pedestrian movement. These enclosure-induced breaks fragment the walkable network and undermine equitable access to key services.

\noindent\textbf{Perceptual Greenness Bias.}
Accounting for gatedness reveals a systematic gap between satellite-derived vegetation and public experience. As summarized in Tab.~\ref{tab:equity_tables}\textbf{(a)}, excluding gated communities (GCs) reduces district-level NDVI, with the largest drops in dense, gated-intensive cores (e.g., Tianhe and Yuexiu), indicating that conventional NDVI aggregation can be sensitive to the spatial footprint of enclosed compounds. This discrepancy is reinforced by street-level perception: Tab.~\ref{tab:equity_tables}\textbf{(b)} shows that the Pearson correlation between Green View Index (GVI) and NDVI decreases under the gatedness-weighted variant (notably in Liwan and Tianhe), whereas the no-gated variant remains comparable or slightly higher. Together, these patterns suggest that a non-trivial portion of NDVI-detected greenery is concentrated within privately controlled gated areas and contributes less to what pedestrians can see or access, motivating enclosure-aware adjustments when using remote sensing to assess environmental equity. The complete district table is provided in the supplementary materials.

\noindent\textbf{Socioeconomic Stratification.}
Shown in Fig.~\ref{fig:expeditionary_maps}\textbf{(right)}, housing price patterns reveal a clear socioeconomic divide between gated and open communities. Gated compounds are concentrated in high-price areas, particularly in Tianhe and other central districts, whereas open communities are more common in older or lower-cost neighborhoods. Even in mixed zones, gated developments consistently align with the upper end of the price spectrum, suggesting that enclosure functions as a value-adding residential amenity and a marker of exclusivity. These patterns indicate that gating is embedded in the city’s housing market dynamics: newly developed high-end projects tend to adopt enclosed forms, while older open blocks face slower reinvestment.

\begin{table*}[!t]
\centering
\caption{\textbf{Equity-related accessibility and greenery metrics.}
\textbf{(a)} District-level NDVI statistics and the estimated bias after excluding gated communities (GCs).
\textbf{(b)} District-wise Pearson correlations between street-level greenery (GVI) and NDVI variants.
\textbf{(c)} Detour accessibility (DI) to representative public-service POIs, quantifying path inflation caused by enclosure.}
\label{tab:equity_tables}
\vspace{-1mm}
\setlength{\tabcolsep}{4pt}
\renewcommand{\arraystretch}{1.05}
\begin{minipage}[t]{0.45\linewidth}
\centering
\resizebox{\linewidth}{!}{
\begin{tabular}{lcccc}
\toprule
\textbf{Metric} & \textbf{Liwan} & \textbf{Tianhe} & \textbf{Yuexiu} & \textbf{Nansha} \\
\midrule
Original NDVI   & 0.1174 & 0.1990 & 0.1412 & 0.1189 \\
NDVI in GCs     & 0.1451 & 0.1635 & 0.1420 & 0.1757 \\
Gated Ratio     & 0.0730 & 0.0874 & 0.0877 & 0.0153 \\
NDVI (No GCs)   & 0.1088 & 0.1816 & 0.1288 & 0.1170 \\
Reduction       & 0.0086 & 0.0174 & 0.0124 & 0.0018 \\
\bottomrule
\end{tabular}
}
\vspace{1mm}

{\footnotesize \textbf{(a)} NDVI by district; \emph{Reduction} denotes the drop by GCs.}
\end{minipage}%
\hfill
\begin{minipage}[t]{0.45\linewidth}
\centering
\resizebox{\linewidth}{!}{
\begin{tabular}{lccc}
\toprule
\textbf{Correlation} & \textbf{Liwan} & \textbf{Tianhe} & \textbf{Yuexiu} \\
\midrule
GVI vs. Raw      & 0.419 & 0.325 & 0.586 \\
GVI vs. No-Gated & 0.405 & 0.332 & 0.588 \\
GVI vs. Weighted & 0.295 & 0.250 & 0.499 \\
\bottomrule
\end{tabular}
}
\vspace{1mm}

{\footnotesize \textbf{(b)} Pearson $r$ between GVI and NDVI variants (\emph{Raw}, \emph{No-Gated}, \emph{Weighted}).}
\end{minipage}

\vspace{2mm}

\begin{minipage}[t]{0.9\linewidth}
\centering
\resizebox{0.9\linewidth}{!}{
\begin{tabular}{lcccc}
\toprule
\textbf{Point of Interest (POI)} & \textbf{Avg. DI} & \textbf{Max DI} & \textbf{Extra Dist. (m)} & \textbf{Blocked Ratio (DI$>$1.05)} \\
\midrule
Taikoo Hui (Commercial Center) & 1.164 & 1.477 & 225.07 & 70.0\% \\
Tianhe Women and Children’s Hospital & 1.102 & 1.239 & 149.93 & 75.0\% \\
\bottomrule
\end{tabular}
}
\vspace{1mm}

{\footnotesize \textbf{(c)} Detour index (DI) and additional travel distance induced by enclosure; \emph{Blocked Ratio} is the share of OD pairs with DI$>$1.05.}
\end{minipage}
\vspace{-10pt}
\end{table*}

\section{Conclusion and Future Work}
\vspace{-2mm}
We presented MCGC, a vision-centric multimodal framework that integrates remote sensing imagery, Chinese metadata, and structured features for recognizing Chinese \textit{fengbi xiaoqu}. The model couples interior and exterior visual streams with cross-modal fusion, achieves strong and interpretable gains over baselines, and supports a metropolitan-scale enclosure map for the Greater Bay Area. Exploratory analyses suggest systematic links between enclosed residential morphology and greenery, pedestrian accessibility, and housing price, providing a practical tool for enclosure-aware urban analytics and computer vision for social good.

\noindent\textbf{Limitations and Ethics.}
Our gated/open definition is locally grounded in the Chinese urban-morphology context and should not be treated as a universal global taxonomy. Broader use requires local definitions, relabeling, and expert validation. In addition, boundary-conditioned modeling still depends on meaningful AOI localization and does not explicitly model all relevant structural cues, such as entrance density, access-control devices, and fine-grained road connectivity. Because residential enclosure labels can be misused for surveillance, geofencing, targeting, or community profiling, Supplementary Sec. E.2 details our tiered release, controlled-access, and DUA restrictions.

\noindent\textbf{Future Work.}
We will expand the benchmark with broader coverage and refined labeling protocols, extend evaluation to cross-city and cross-cultural settings with local taxonomies, domain adaptation and calibration. We will also maintain the dataset/benchmark repository at \dataseturl~as a living release with updated documentation and expanded non-sensitive datasets.
\newpage

%
%
\bibliographystyle{splncs04}
\bibliography{eccv}
\end{document}


\title{Supplementary Material for ``Urban Boundaries, Social Barriers''}
\titlerunning{Supplementary Material}
\author{Minwei Zhao\inst{1}\orcidlink{0000-0002-6380-5426}\thanks{These authors contributed equally.} \and
Weiming Zhang\inst{1}\orcidlink{0009-0003-2278}\textsuperscript{*} \and
Jiawang Du\inst{1}\orcidlink{0000-0002-1334-4158} \and
Qiming Liu\inst{2,1}\orcidlink{0009-0006-8421-0305} \and
Weiming Zhuang\inst{3}\orcidlink{0000-0001-8243-7772} \and
Pei Nie\inst{4}\orcidlink{0000-0003-2370-2112} \and
Cai Wu\inst{1}\orcidlink{0000-0002-5578-5525}\thanks{Corresponding author.}}
\authorrunning{M.~Zhao et al.}
\institute{The Hong Kong University of Science and Technology (Guangzhou)\\
\email{\{m.zhao,wzhang915,jdu146\}@connect.hkust-gz.edu.cn, caiwu@hkust-gz.edu.cn}
\and
School of Public Administration and Policy, Renmin University of China\\
\email{qimingliu937@ruc.edu.cn}
\and
Sony AI\\
\email{weiming.zhuang@sony.com}
\and
University of South China\\
\email{niepei@usc.edu.cn}}
\maketitle
\setcounter{page}{1}
\renewcommand{\thesection}{S\arabic{section}}
\renewcommand{\thesubsection}{S\arabic{section}.\arabic{subsection}}
\renewcommand{\thesubsubsection}{S\arabic{section}.\arabic{subsection}.\arabic{subsubsection}}
\renewcommand{\theequation}{S\arabic{equation}}

\renewcommand{\thefigure}{A\arabic{figure}}
\renewcommand{\thetable}{A\arabic{table}}
\setcounter{figure}{0}
\setcounter{table}{0}
\section{Greater Bay Area Dataset and Exploratory Analysis}
\label{Supplement.A}
\subsection{Dataset Statistics by City}
\label{Supplement.A1}
\begin{figure}[!h]
    \centering
    \includegraphics[width=\linewidth]{all GBA.jpg}
    \caption{Spatial distribution of gated(grey) and open(red) communities across the mainland Greater Bay Area. Derived from our \textit{MCGC} classification, the benchmark includes 37,444 AOIs labeled by gating status.}
    \label{fig:gba_full}
\end{figure}

\begin{figure}[!h]
    \centering
    \includegraphics[width=\linewidth]{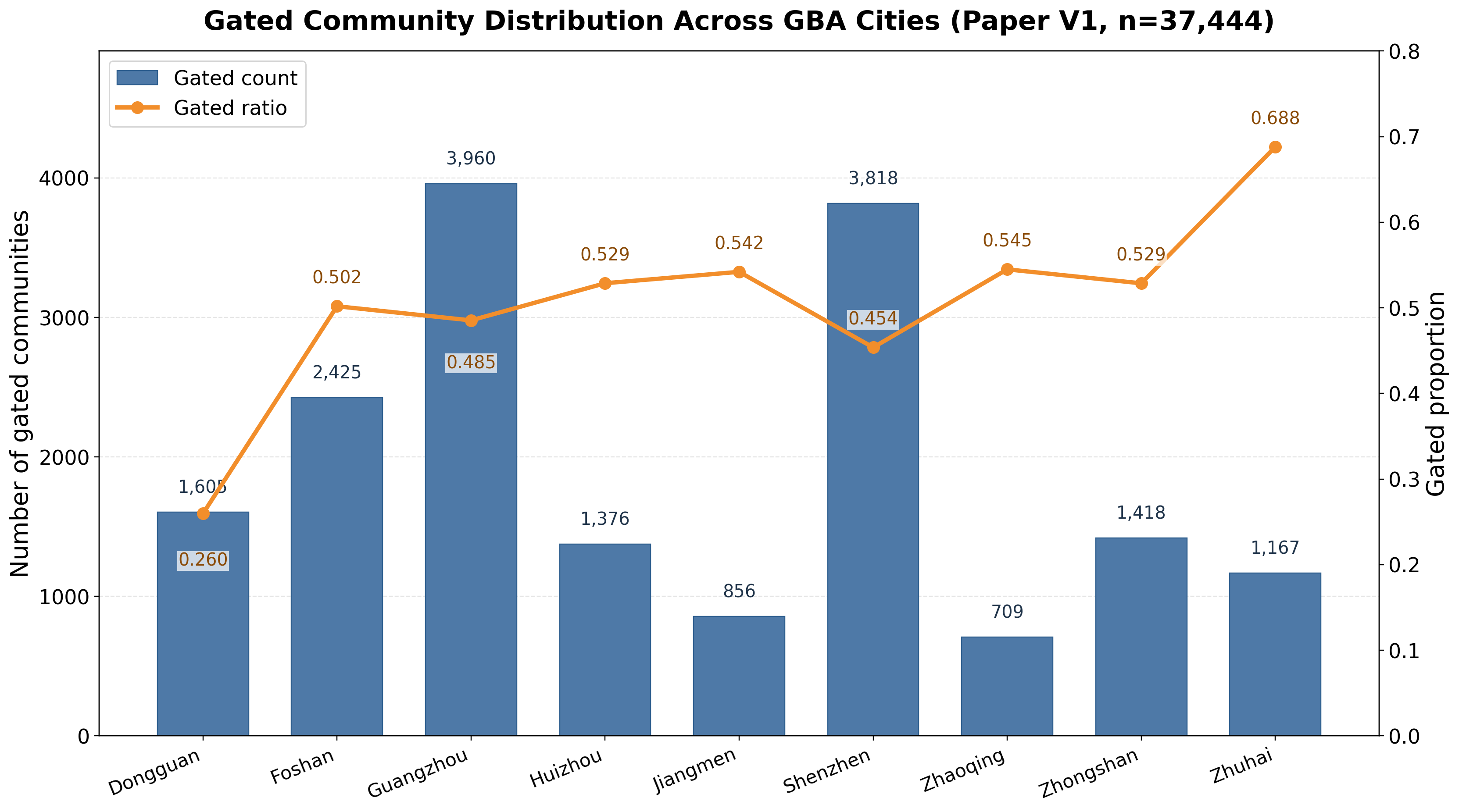}
    \caption{Gated community distribution across Greater Bay Area cities. Bar chart shows absolute number; line chart shows gated proportion.}
    \label{fig:gated_stats}
\end{figure}
Figure~\ref{fig:gba_full} provides an overview of the full benchmark distribution across the Greater Bay Area (GBA), covering residential community AOIs identified by our \texttt{MCGC} classification framework. The benchmark contains 37,444 communities, each labeled as either gated or open. The spatial coverage encompasses nine major mainland GBA cities and reveals distinctive urban morphological patterns, offering a foundation for subsequent analysis of spatial gating and its urban implications.

\begin{figure*}[!t]
    \centering
    \includegraphics[width=1\linewidth]{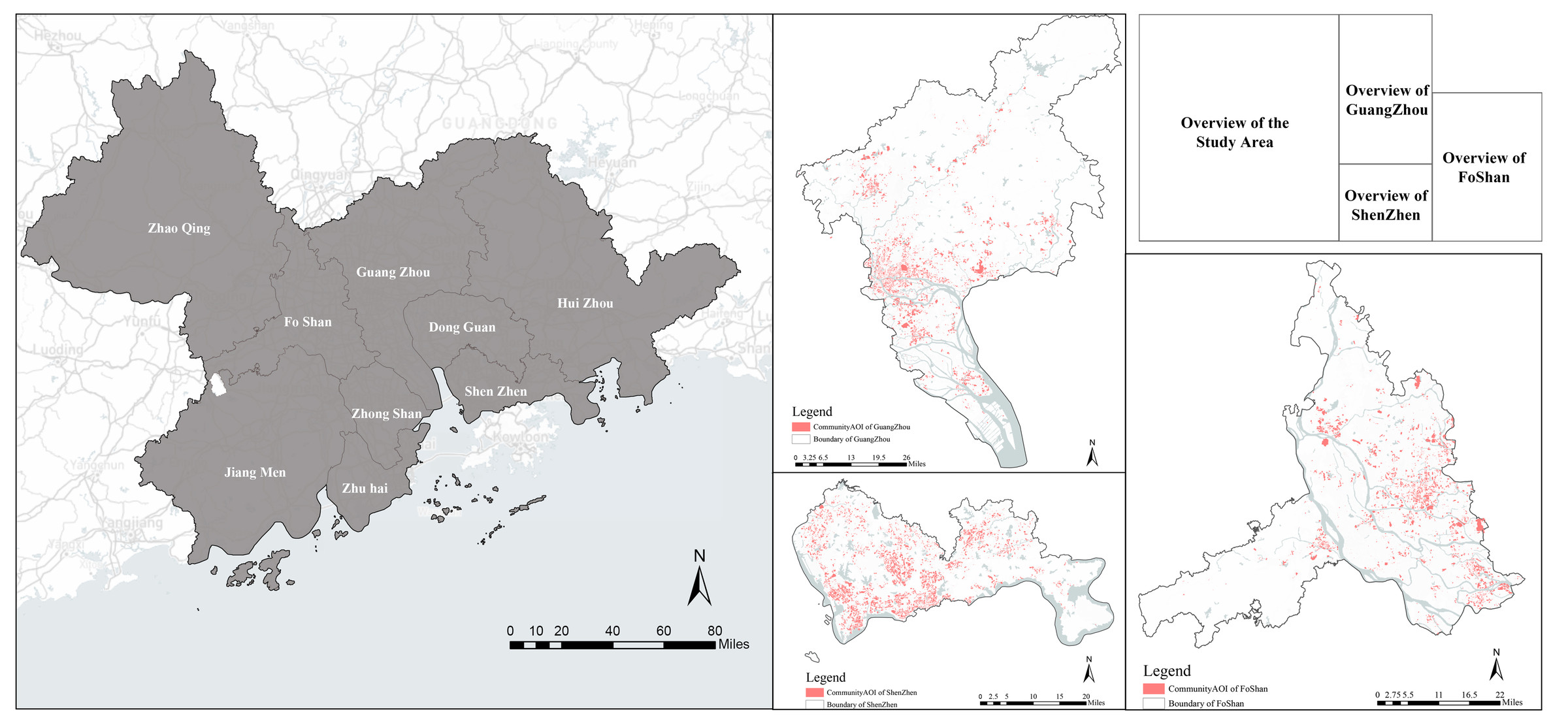}
    \caption{Spatial distribution of  communities across three representative cities-Guangzhou, Shenzhen, and Foshan. The left panel shows the location of these cities within the Greater Bay Area, while the right panels provide detailed local maps of community gating patterns.}
    \label{fig:overreview}
\end{figure*}

Figure~\ref{fig:overreview} primarily provides an overview of the study area. This research focuses on nine cities within the Greater Bay Area (excluding Macao and Hong Kong) and collects relevant data. Using Guangzhou City as an example for dataset training, the study extends its prediction of gated community types to the remaining eight cities. To investigate the spatial distribution differences between gated and open communities, the study selects three typical cities—Guangzhou, Shenzhen, and Foshan—for further spatial analysis after the prediction on the Greater Bay Area dataset.

Figure~\ref{fig:gated_stats} presents the absolute count (blue bars) and proportion (orange line) of gated communities in nine major cities using the stable 37,444-AOI paper benchmark. Guangzhou and Shenzhen exhibit the highest counts, with 3,960 and 3,818 gated communities respectively. However, the gated ratios vary, with Zhuhai (0.688), Jiangmen (0.542), and Zhaoqing (0.545) showing higher relative prevalence despite lower absolute numbers. This divergence reflects differences in planning regimes, development models, and historical urbanization patterns across cities.

These disparities suggest that gating is not merely a function of urban scale or population but is shaped by divergent urban development strategies. Smaller cities such as Zhuhai, Zhaoqing, and Jiangmen can exhibit high relative prevalence, potentially reflecting local development histories and real-estate-led growth. The gated ratios range from 0.260 to 0.688 across cities, indicating substantial spatial heterogeneity in privatized residential configurations that may influence urban equity, green space accessibility, and mobility.

\subsection{Spatial Statistical Analysis of Communities in Three Typical Cities}

\begin{figure*}[!h]
    \centering
    \includegraphics[width=\linewidth]{Spatial Distribution of Ungated and Gated Communities in Three Typical Cities.jpg}
    \caption{Spatial distribution of gated (red) and open (blue) communities in three representative cities of the Greater Bay Area—Guangzhou, Shenzhen, and Foshan. The maps highlight intra-city differences in gating patterns and community morphology.}
    \label{fig:gated_distribution_threecities}
\end{figure*}

\begin{figure*} [!h]
    \centering
    \includegraphics[width=1\linewidth]{Bivariate Spatial Autocorrelation Analysis.jpg}
    \caption{Bivariate Spatial Autocorrelation Analysis: LISA and Significant Maps }
    \label{fig:moranI}
\end{figure*}

By visualizing the spatial distribution of open and gated communities through the prediction results (Figure~\ref{fig:gated_distribution_threecities}), the study found that in the three cities (Guangzhou, Shenzhen, and Foshan), the number of gated communities generally exceeded that of open communities. Open communities were predominantly located in urban central areas or old urban districts, while gated communities were more dispersed in their distribution.

To further investigate the spatial association between different community types (Figure~\ref{fig:moranI}), the study aggregated the number of gated and open communities into 200m × 200m grid cells and employed bivariate local Moran's I to examine their spatial autocorrelation. The bivariate Moran's I indices for Guangzhou, Shenzhen, and Foshan were 0.154, 0.124, and 0.102, respectively, indicating spatial clustering of both open and gated communities in these three cities. The Local Indicators of Spatial Association (LISA) maps of the three cities further revealed the forms of community spatial clustering:
1. High-high (open vs. gated) community clusters were mainly concentrated in urban core areas.
2. Low-high community clusters were widely distributed and represented the predominant form of community clustering.
3. High-low and low-low community clusters were sparsely distributed.
4. Other areas exhibited insignificant clustering patterns, constrained by the natural environment of the cities.

Constrained by urban development models and limitations in productivity, early urban construction in the Guangdong–Hong Kong–Macao Greater Bay Area was often relatively concentrated, predominantly following a pattern of high-density buildings and high-density road networks. Similar to traditional Chinese residential lifestyles, community interactions were primarily characterized by visits among neighbors, and thus community gates and access restrictions were typically not installed or were less frequently implemented. Since 2000, as China has vigorously pursued urban expansion and the real estate industry has flourished, residential construction has increasingly focused on the quality of neighborhood environments, spatial privacy, and site selection. Gated residential communities are usually uniformly constructed by developers to reduce the costs of municipal supporting facilities. In contrast, due to fragmented property rights and high demolition costs, old urban areas cannot be entirely transformed into gated communities, resulting in the preservation of a large number of open communities. This has led to the spatial characteristics of the Greater Bay Area cities, where gated communities are numerous and widely distributed, while open communities are fewer in number and more concentrated in their distribution. This result reveals the general patterns of community distribution types, providing data support for subsequent urban renewal and transformation.

\section{Dataset Construction and Annotation Protocol}
\label{Supp:Annotation}
\subsection{Collection Pipeline and Multimodal Composition}
GBA-GCs is constructed from residential AOIs in the nine mainland Greater Bay Area cities. For each AOI, we align three modalities: (i) a licensed map-provider AOI polygon used for spatial cropping and area computation; (ii) 0.8\,m remote-sensing imagery cropped with a 50\,m context buffer to preserve walls, entrances, street edges, and interior--exterior transitions; and (iii) textual and structured attributes, including provider metadata, floor-area ratio, and POI statistics when available. The benchmark task is AOI-level gated/open recognition under heterogeneous modality availability. Public release files expose only anonymized row identifiers, centroids, labels, and coarse derived attributes; full polygons, names, addresses, imagery, and provider metadata are controlled as described in Supplementary Sec. E.2.

\subsection{Operational Label Definition}
We use a locally grounded definition of Chinese \textit{fengbi xiaoqu}. A gated AOI is a residential compound with a continuous or near-continuous perimeter, limited or controlled entrances, and partial discontinuity between internal circulation and the surrounding public street network. An open AOI is publicly permeable, lacks compound-level access control, or is integrated with surrounding streets. Annotators cross-check satellite imagery, street-view evidence when available, metadata, road connectivity, and planning records. Ambiguous cases include partial gating, mixed-use villages, weak or occluded entrances, redevelopment sites, and conflicting source evidence.

\subsection{Annotators, Redundancy, and Adjudication}
Eighteen annotators with MSc-level or higher training in urban planning, geography, architecture, or related spatial-analysis fields participated in the annotation process. Each AOI in the controlled Guangzhou evaluation subset receives at least three redundant judgments. Disagreements and low-confidence cases are escalated to senior adjudication by PhD-level researchers or practitioners using the same operational definition. This procedure yields 94\% duplicate-set agreement and Cohen's $\kappa=0.85$ overall, with per-subset agreement in the range of $\kappa=0.81$--$0.90$. The final Guangzhou evaluation set contains 5,268 labeled AOIs: 2,605 gated and 2,663 open.

\subsection{External Diagnostic Verification}
Hong Kong, Singapore, and London are used only as diagnostic transfer settings, not as evidence for a universal gated/open taxonomy. London and Singapore labels are verified by at least three local annotators with MSc-level or higher training and at least three years of local residence or research experience. These diagnostic checks are reported separately from the mainland GBA benchmark and should be interpreted under local morphology and data-quality constraints.

\renewcommand{\thefigure}{B\arabic{figure}}
\renewcommand{\thetable}{B\arabic{table}}
\setcounter{figure}{0}
\setcounter{table}{0}
\section{Social Equity and Spatial Analysis Methodologies}
\label{Supp:B}
\setcounter{equation}{0}
\subsection*{B.1 Microscale Spatiotemporal Analyses: The Case of Guangzhou}
To further demonstrate the extensibility of our dataset, we conduct microscale analyses in Guangzhou’s central urban area. These case studies integrate multiple spatial data sources to examine how enclosure boundaries influence accessibility, environmental perception, and socioeconomic structure. The analyses exemplify how enclosure-aware urban data can support interdisciplinary research on spatial equity, environmental sustainability, and inclusive city design. 

\subsection*{B.2 Temporal Emergence of Gated Communities} We intersect community footprints with multi-temporal urban expansion layers to analyze how gated communities emerged across four developmental phases: pre-1985, 1985–2005, post-2005, and outer non-central areas. This temporal segmentation provides a framework for tracing enclosure evolution alongside urbanization dynamics. 

\subsection*{B.3 Quantifying Greenery Perception Mismatch} To assess the environmental implications of gating, we implement a spatial overlay that estimates the vegetation bias introduced by private enclosures. For each district $j$, the gated area ratio is defined as $r_j = A_j^{\text{gated}} / A_j$, and the gated-excluded NDVI as \begin{equation}     \text{NDVI}_{\text{nogated}} = \text{NDVI}_{\text{original}} (1 - r_j). \end{equation} The difference $\Delta_j = \text{NDVI}_{\text{original}} - \text{NDVI}_{\text{nogated}}$ provides a proxy for the masking effect of gated communities on perceived greenery, illustrating how enclosure boundaries can distort environmental indicators. 

\subsection*{B.4 Simulating Gating-Induced Detours} To model the mobility constraints imposed by physical enclosure, we design a geometry-based simulation independent of road networks. Synthetic origins are uniformly sampled within 3 km buffers centered on two representative points of interest, namely \textit{Taikoo Hui} and \textit{Tianhe District Women and Children’s Hospital}. Each agent computes both a straight path and a gated-aware route using a greedy step algorithm that adjusts direction whenever a barrier is encountered. The detour index for agent $i$ is defined as \begin{equation} \mathrm{DI}_i = \frac{\tilde{d}_i}{d_i}, \quad \mathrm{DI}_i \geq 1, \end{equation} and the overall mean is \begin{equation} \overline{\mathrm{DI}} = \frac{1}{N}\sum_{i=1}^{N}\mathrm{DI}_i, \end{equation} offering a standardized estimate of enclosure-induced impedance on pedestrian movement. 

\subsection*{B.5 Socioeconomic Correlates of Gating} To explore potential socioeconomic correlates, we integrate housing price data across four central districts (Tianhe, Yuexiu, Liwan, and Haizhu) and classify each community as either gated or open. Price gradients are visualized using continuous color maps, allowing analysis of the spatial co-location between gated morphology and socioeconomic stratification. 
Collectively, these analyses serve as preliminary yet illustrative examples of how our dataset and framework can support scalable, reproducible, and socially meaningful studies. They highlight the potential of enclosure-aware urban computing to bridge computer vision, spatial analysis, and urban equity research.

\renewcommand{\thefigure}{C\arabic{figure}}
\renewcommand{\thetable}{C\arabic{table}}
\setcounter{figure}{0}
\setcounter{table}{0}
\section{Exploratory Analyses on Spatial Enclosure and Its Urban Implications}
\subsection{Gated Community-induced Detour Effects on Pedestrian Accessibility}
\label{Supp:C1}
\begin{figure}[!h]
    \centering
    \includegraphics[width=0.8\linewidth]{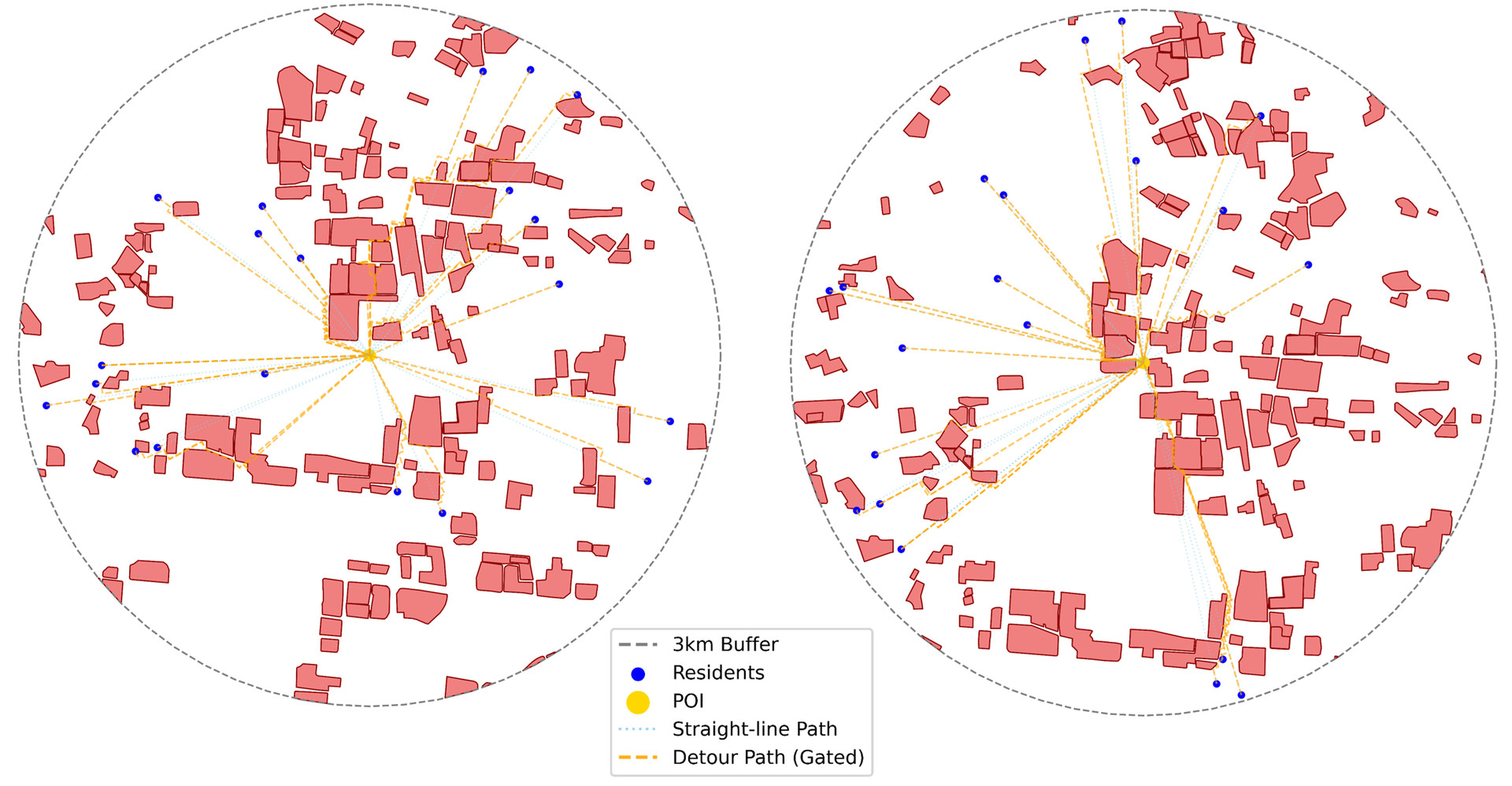}
    \caption{Simulated detour paths due to gated community barriers. Left: access to Taikoo Hui; Right: access to Tianhe Women and Children’s Hospital. Orange lines represent detour-aware pedestrian routes avoiding gated perimeters (red), contrasting with ideal straight-line access (blue).}
    \label{fig:detour}
\end{figure}
Figure~\ref{fig:detour} illustrates simulated pedestrian paths from randomly sampled residents within a 3 km radius of two key POIs. Each subplot contrasts ideal straight-line trajectories with detour-aware routes that avoid gated community boundaries.

Results reveal substantial accessibility penalties. The average Detour Index (DI) reaches 1.133, peaking at 1.477, with average additional travel distances exceeding 100 meters. Notably, over 70\% of residents exhibit DI values above 1.05, indicating widespread enclosure-induced distortion.

These findings underscore how gated morphologies operate as impermeable pedestrian barriers—reducing connectivity and introducing structural inequality in service access. Unlike natural obstacles, gated boundaries reflect socio-spatial privatization, reinforcing exclusionary dynamics in the urban fabric. This effect is especially acute in dense urban cores like Tianhe, where overlapping enclosures fragment walkable paths and channel movement into constrained corridors.
\begin{figure}[!t]
    \centering
    \includegraphics[width=0.8\linewidth]{housing price.jpg}
    \caption{
    Spatial distribution of housing prices for gated (circles) and open (triangles) communities across Guangzhou’s four central districts (Tianhe, Yuexiu, Liwan, Haizhu). 
    }
    \vspace{-10pt}
    \label{fig:housing_price_app}
\end{figure}
\renewcommand{\thefigure}{C.\arabic{figure}}
\setcounter{figure}{0}
\renewcommand{\thetable}{C.\arabic{table}}
\setcounter{table}{0}
\label{Supp:C}
\begin{table*}[!t]
\centering
\caption{District-level NDVI and reduction after gated community exclusion}
\label{tab:ndvi_gated_effect}
\begin{tabular}{lccccc}
\toprule
\textbf{District} & \textbf{Original NDVI} & \textbf{NDVI in GCs} & \textbf{Gated Ratio} & \textbf{NDVI (No GCs)} & \textbf{Reduction} \\
\midrule
Huangpu     & 0.2398 & 0.1925 & 0.0287 & 0.2329 & 0.0069 \\
Huadu       & 0.2394 & 0.1945 & 0.0226 & 0.2340 & 0.0054 \\
Liwan       & 0.1174 & 0.1451 & 0.0730 & 0.1088 & 0.0086 \\
Zengcheng   & 0.2980 & 0.1867 & 0.0174 & 0.2928 & 0.0052 \\
Tianhe      & 0.1990 & 0.1635 & 0.0874 & 0.1816 & 0.0174 \\
Haizhu      & 0.1444 & 0.1543 & 0.0834 & 0.1323 & 0.0120 \\
Conghua     & 0.3012 & 0.2265 & 0.0048 & 0.2998 & 0.0015 \\
Yuexiu      & 0.1412 & 0.1420 & 0.0877 & 0.1288 & 0.0124 \\
Baiyun      & 0.2195 & 0.1760 & 0.0232 & 0.2144 & 0.0051 \\
Panyu       & 0.1604 & 0.1801 & 0.0608 & 0.1507 & 0.0098 \\
Nansha      & 0.1189 & 0.1757 & 0.0153 & 0.1170 & 0.0018 \\
\bottomrule
\end{tabular}
\end{table*}
\subsection{Disturbance of gated communities on regional NDVI}
\label{Supp:C2}
To examine how gated communities distort urban greenness assessments, we analyzed NDVI values both within and outside gated areas across districts in Guangzhou. Results reveal that NDVI inside gated compounds is generally lower, and their spatial footprint leads to an overestimation of district-level vegetation coverage when not properly accounted for.
Table~\ref{tab:ndvi_gated_effect} summarizes the district-level impact of gated communities on NDVI measurements across eleven administrative units in Guangzhou. A consistent pattern emerges: in nearly all districts, the NDVI within gated communities is substantially lower than the overall district-level average. For example, in Huangpu, NDVI drops from 0.2398 to 0.1925 when focusing solely on gated areas, and in Zengcheng—a district with the highest baseline NDVI—the gated subset shows a significant reduction to 0.1867. This suggests that gated neighborhoods may have denser built environments, reduced greenery allocation, or lower public greening effort due to privatized land control.

The results further indicate that districts with higher \textit{Gated Area Ratios} tend to exhibit more severe NDVI distortions. Tianhe and Yuexiu—two central urban districts with gated coverage above 8\%—show the largest NDVI reductions (0.0174 and 0.0124, respectively), underscoring how spatial enclosure contributes to the erasure of vegetative perception in high-density urban cores. Interestingly, even in peripheral districts like Conghua and Zengcheng, where gated coverage is below 2\%, gated NDVI is significantly lower than the district mean, suggesting potential ecological fragmentation effects.

These findings highlight the necessity of accounting for spatial enclosure in urban vegetation analysis. Without disaggregating the gated component, conventional NDVI aggregation may yield misleading conclusions, especially in cities with widespread privatized enclaves. 

\begin{table}[!h]
\centering
\setlength{\tabcolsep}{4pt}
\caption{Pearson correlations between GVI and NDVI variants by district}
\label{tab:gvi_ndvi_corr}
\begin{tabular}{lccc}
\toprule
\textbf{District} 
& \makecell[c]{GVI\\vs. Raw} 
& \makecell[c]{GVI\\vs. No-Gated} 
& \makecell[c]{GVI\\vs. Weighted} \\
\midrule
Huangpu  & 0.266 & 0.269 & 0.263 \\
Liwan    & 0.419 & 0.405 & 0.295 \\
Tianhe   & 0.325 & 0.332 & 0.250 \\
Haizhu   & 0.360 & 0.358 & 0.260 \\
Yuexiu   & 0.586 & 0.588 & 0.499 \\
Baiyun   & 0.372 & 0.368 & 0.280 \\
Panyu    & 0.448 & 0.450 & 0.299 \\
\bottomrule
\end{tabular}
\caption*{\footnotesize Note: GVI = Green View Index; Raw, No-Gated, and Weighted refer to different NDVI variants.}
\vspace{-10pt}
\end{table}

Vegetation assessment in urban areas often relies on satellite-derived metrics like the Normalized Difference Vegetation Index (NDVI). However, such indicators may be distorted by spatial enclosures that limit visual and physical access to green spaces. To evaluate this discrepancy, we compared the Pearson correlations between street-level Green View Index (GVI)—a proxy for perceived greenery—and three NDVI variants: (1) raw NDVI, (2) NDVI with gated areas excluded (No-Gated), and (3) NDVI weighted by gated community presence (Weighted).

As shown in Table~\ref{tab:gvi_ndvi_corr}, GVI exhibits moderate correlations with raw NDVI in most districts, yet these correlations decrease substantially under the Weighted variant in nearly all cases. Notably, in Liwan and Tianhe—two densely built districts with a high proportion of gated communities—the correlation drops from 0.419 to 0.295 and from 0.325 to 0.250, respectively. This consistent pattern suggests that including or overweighting gated green space introduces a perceptual mismatch between satellite-based greenness and ground-level human experience. In contrast, the No-Gated NDVI variant marginally improves or stabilizes the GVI correlation, implying that gated vegetation—although present in spectral data—is not functionally or visually accessible to the public.

These results underscore a critical limitation in conventional NDVI-based urban greenness evaluations: they may overestimate public greenery in cities where large portions of vegetation are sequestered within privately controlled gated compounds. Our findings thus reinforce the need for incorporating visibility and access-based adjustments when using remote sensing data to assess environmental equity and urban livability.
\begin{figure*}[!t]
    \vspace{-5pt}
    \centering
    \includegraphics[width=0.85\linewidth]{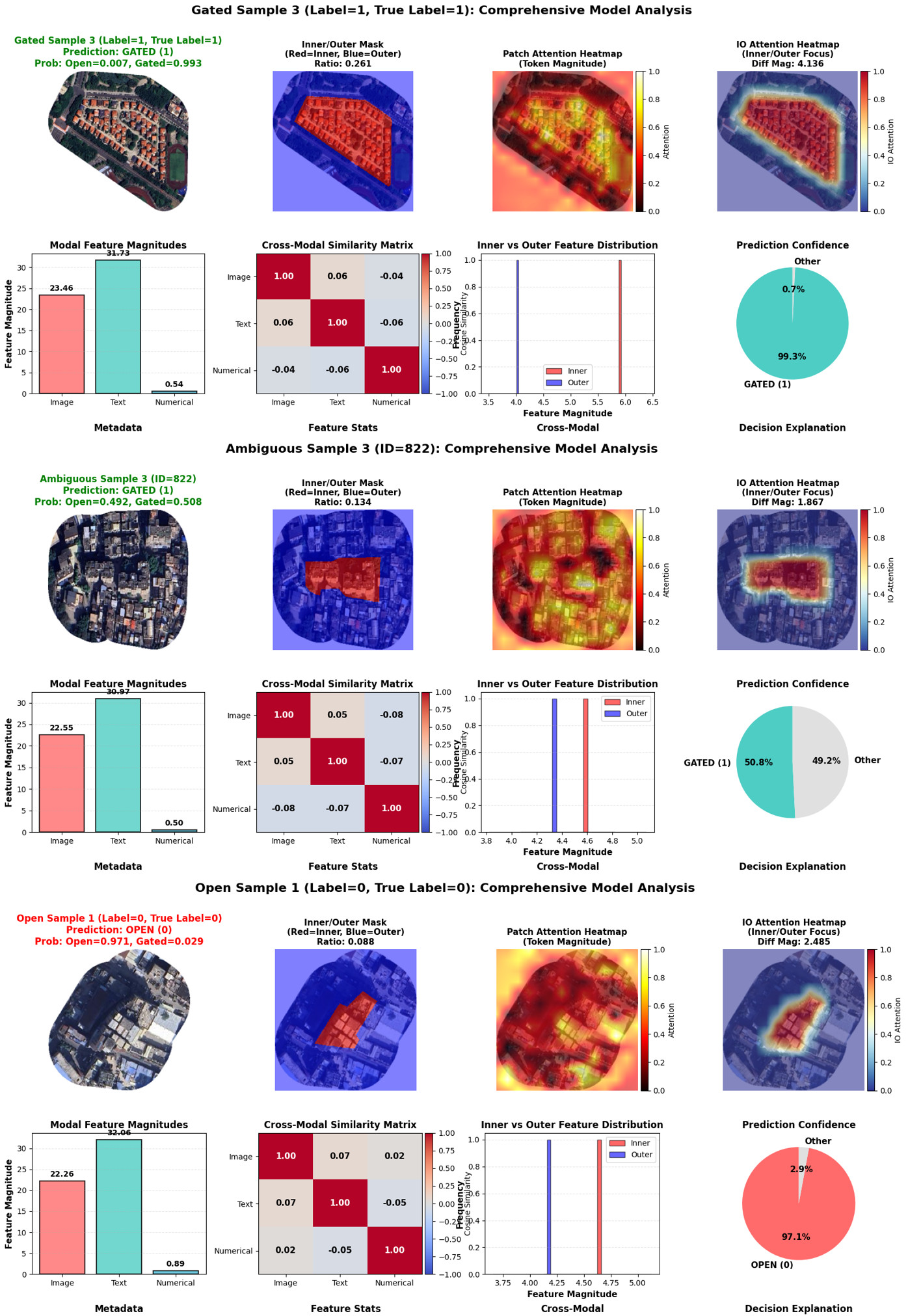}
    \caption{Representative model interpretation cases. Each row shows the original AOI image, attention heatmaps (patch-level and inner–outer focus), and confidence distributions for gated and ambiguous samples. High-confidence cases emphasize enclosure edges, while ambiguous ones display diffuse attention and balanced confidence, reflecting genuine spatial uncertainty in human annotation.}
    \label{fig:case_study_supp_1}
\end{figure*}
\subsection{Socioeconomic Stratification Associated with Gated Communities}
\label{Supp:C3}
Figure~\ref{fig:housing_price_app} maps housing prices across Guangzhou’s four central districts—differentiating between gated (circles) and open (triangles) communities. A distinct spatial pattern emerges: gated communities cluster in high-value zones (¥80,000–¥100,000+/m\textsuperscript{2}), often marked in red and orange, while open communities appear more frequently in lower-priced areas, particularly in southern Liwan and Haizhu.

Even within localized neighborhoods where both types coexist, gated communities consistently exhibit higher housing prices than their open counterparts. This suggests a strong correlation between community enclosure and property value, with gating potentially acting as a spatial proxy for social status and exclusivity.

These findings underscore how gating contributes to spatial and socioeconomic segregation. The premium attached to gated communities may reinforce urban inequality—both in access and affordability—and hinder social integration. As cities like Guangzhou densify and redevelop, the unchecked expansion of high-priced gated enclaves could pose challenges to inclusive urban growth and equitable housing provision.

\renewcommand{\thefigure}{D\arabic{figure}}
\renewcommand{\thetable}{D\arabic{table}}
\setcounter{figure}{0}
\setcounter{table}{0}
\section{Code, Data, and Implementation Details}
\label{Supp:D}
\begin{table*}[t]
  \centering
  \caption{Baseline configuration summary. ``Fusion'' denotes how modalities are combined; ``FT'' indicates whether pretrained encoders are fine-tuned; ``LoRA'' marks where low-rank adapters are applied. ``Concat'' means simple concatenation of pooled embeddings. ``Cross-attn'' refers to the multi-stage cross-attention used in MCGC. LoRA is injected into the query/value projections of the DINOv3 image encoder.}
  \label{tab:baseline-config}
  \resizebox{\textwidth}{!}{%
    \begin{tabular}{lcccc}
      \toprule
      Model & Modalities & Fusion & FT & LoRA \\
      \midrule
      Text-only & Text & Pooling + Linear & Yes & No \\
      Image-only & Image & Pooling + Linear & Yes & Yes (Q/V) \\
      Numeric-only & Numeric & Linear & Yes & No \\
      Text + Image (naive) & Text, Image & Concat & Yes & Yes (image) \\
      Text + Numeric (naive) & Text, Numeric & Concat & Yes & No \\
      Image + Numeric (naive) & Image, Numeric & Concat & Yes & Yes (image) \\
      MCGC (ours) & Text, Image, Numeric & Multi-stage cross-attn + Concat & Yes & Yes (image) \\
      \bottomrule
    \end{tabular}%
  }
\end{table*}

\subsection{Technical Details}
\label{Supp:D1}
\subsubsection{LoRA Configuration}
We integrate LoRA into the DINOv3 image encoder to enable parameter-efficient fine-tuning. The adapters are injected into the query and value projection matrices of each multi-head self-attention block. Each low-rank update is decomposed as \( \Delta W = BA \), where \( B \in \mathbb{R}^{d \times r} \), \( A \in \mathbb{R}^{r \times d} \), and \( r \) is the rank. We set \( r=4 \) and apply a scaling factor \( \alpha \) such that the effective update is \( \frac{\alpha}{r} BA \); the original pretrained weights remain frozen. This allows adaptation to gatedness-specific visual patterns with a small parameter overhead. (If LoRA is applied to other modules, e.g., the text encoder, specify here similarly.)

\subsubsection{Baseline Configuration}
Table~\ref{tab:baseline-config} summarizes the key differences among our compared models, ensuring a fair comparison by aligning training hyperparameters unless otherwise noted.

All baselines share the same optimization hyperparameters (learning rate, weight decay, dropout rate, number of epochs, early stopping patience) unless explicitly stated, to isolate architectural effects.

\subsubsection{Cross-Validation Protocol}
We conduct five-fold cross-validation by randomly partitioning the dataset into five disjoint subsets with a fixed seed. Each fold uses 80\% for training and 20\% for validation. While this random split may admit some spatial dependence across folds, we leave more conservative geographically-aware splitting to future work. Early stopping is applied based on validation AUC with a patience of 5 epochs. Reported results are the mean and standard deviation over the folds. Unless otherwise specified, each reported result represents the mean performance over five random splits. To ensure fair comparison, all train–test partitions are generated using stratified sampling, maintaining an approximately equal proportion of \textit{gated} and \textit{open} communities in each fold.

\subsection*{D.2 Case Study: Model Interpretation and Visual Analysis}
\label{Supp:D2}
To further interpret MCGC’s multimodal decision process, we visualize representative samples from the Guangzhou dataset in Figure~\ref{fig:case_study_supp_1}. Each row corresponds to a gated or ambiguous case, illustrating image-level attention, inner–outer (IO) attention focusing, and prediction confidence distributions. These visual diagnostics provide insight into how the model encodes enclosure cues and handles uncertain boundary conditions.

In high-confidence gated samples (top), both patch-level and IO attention maps concentrate along enclosure edges and entry points, reflecting strong correspondence with human-interpretable morphological boundaries. In contrast, ambiguous samples (bottom) show weaker and more spatially diffused attention, often arising from visually mixed or partially open residential forms. Notably, these ambiguous cases were also rated as uncertain by human annotators, suggesting that the model’s intermediate uncertainty aligns with genuine perceptual ambiguity rather than prediction noise.

Overall, these qualitative analyses complement the quantitative results in Section~4.3, indicating that MCGC captures enclosure-relevant visual semantics and expresses uncertainty consistently with human judgment.

\subsection*{D.3 Case Study: IO Fusion Strategy Analysis}
\label{Supp:D3}

\begin{table*}[!h]
\centering
\setlength{\tabcolsep}{4pt}
\caption{Comparison of the Performance of different integration mechanisms on IO}
\label{tab:IO_Fusion}

\resizebox{\textwidth}{!}{%
\begin{tabular}{l ccccc}
\toprule
\textbf{Different Fusion Strategies} 
& \textbf{Acc} 
& \textbf{Prec} 
& \textbf{Rec} 
& \textbf{F1} 
& \textbf{AUC}  \\
\midrule
Without IO  & 0.8383 ± 0.0310 & 0.8541 ± 0.0465 & 0.8150 ± 0.0500 & 0.8326 ± 0.0330 & 0.9096 ± 0.0283 \\
Direct IO Fusion    & 0.8109 ± 0.0190 & 0.8169 ± 0.0325 & 0.7992 ± 0.0378 & 0.8069 ± 0.0195 & 0.8892 ± 0.0182 \\
Separate IO with Multimodal Fusion   & 0.8015 ± 0.0231 & 0.8409 ± 0.0272 & 0.7386 ± 0.0332 & 0.7861 ± 0.0260 & 0.8814 ± 0.0223 \\
IO-enhanced Image Feature  & 0.8305 ± 0.0270 & 0.8355 ± 0.0486 & 0.8111 ± 0.0303 & 0.8218 ± 0.0231 & 0.8972 ± 0.0275 \\
IO-Difference Enhanced Fusion  & 0.8176 ± 0.0245 & 0.8383 ± 0.0368 & 0.7846 ± 0.0404 & 0.8095 ± 0.0263 & 0.8912 ± 0.0204 \\
IO with BCA (\textbf{ours})    & 0.8531 ± 0.0376 & 0.8694 ± 0.0518 & 0.8296 ± 0.0333 & 0.8485 ± 0.0372 & 0.9173 ± 0.0393 \\
\bottomrule
\end{tabular}%
}
\vspace{-10pt}
\end{table*}

Table~\ref{tab:IO_Fusion} reports a detailed comparison of alternative ways of incorporating inner-outer (IO) cues into our multimodal CCF architecture.

\noindent\textbf{Without IO.}
This baseline uses only the three modalities (text, image, and numerical attributes) with the original CCF block, without any IO-related cues. 
Even in this configuration, the model achieves strong performance 
(Acc = 0.8383, F1 = 0.8326, AUC = 0.9096), 
demonstrating that the fundamental tri-modal design is already highly effective.

\noindent\textbf{Direct IO Fusion.}
Instead of directly gating the tri-modal features with IO cues, 
this variant first extracts inner and outer features using the numerical boundary information, 
and applies an independent convolutional transformation to each of them to obtain two IO-aware feature vectors. 
These IO-enhanced features are then incorporated as additional signals during the final fusion stage: 
together with the three modality-specific cross-attended features (image--text, image--numerical, and text--numerical), 
all five feature vectors are re-weighted through a learned fusion head to produce the final representation. 
However, despite introducing IO cues, this direct concatenation-and-gating design still leads to clear performance degradation (e.g., Acc, F1, and AUC all drop compared with the IO-free baseline), 
suggesting that shallow IO transformations without explicit contrast reasoning are insufficient for reliable multimodal fusion.

\noindent\textbf{Separate IO with Multimodal Fusion.}
In this variant, we explicitly separate the IO cues and let the inner and outer features interact with different modalities. 
The inner feature, which captures the structural and functional characteristics within the community boundary, 
replaces the original image representation when performing tri-modal cross-attention with both the text and numerical branches. 
This design allows the model to better associate community interior patterns with the community name (text) and structured attributes (numerical features). 
In contrast, the outer feature is assumed to be more semantically aligned with the textual modality---%
the surrounding environment of a community (e.g., adjacent roads, landscape context) often correlates with its designated name or type. 
Thus, the outer representation only attends to the text encoder to produce an outer--text interaction feature. 
Finally, the three cross-attended features from the inner branch and the outer--text feature are jointly fused through a convolutional weighting head to form the final representation. 
Although this variant recovers some precision compared to Direct IO Fusion, 
its overall recall (0.7386) and F1 (0.7861) remain significantly lower than the IO-free baseline, 
indicating that simply separating the IO streams without modeling their explicit contrast leads to incomplete spatial reasoning and suboptimal multimodal fusion.

\noindent\textbf{IO-enhanced Image Feature.}
This variant aims to enrich the visual backbone with explicit interior--exterior structure before any multimodal interaction occurs. 
Starting from the baseline image tokens, we first use the numerical boundary information to decompose the image representation into an inner token set and an outer token set. 
Each set undergoes its own self-attention operation, enabling the model to capture region-specific spatial patterns---%
the inner region reflecting residential morphology inside the boundary, and the outer region encoding surrounding environmental context. 
To further model the structural relationship between these two regions, we perform a dedicated cross-attention step between the inner and outer tokens, 
allowing each region to exchange complementary cues such as enclosure contrast, boundary consistency, and contextual transitions. 
The resulting IO-refined tokens are then fused back into the original global image representation through a residual integration path, 
ensuring that the enhanced IO-aware information strengthens, rather than overrides, the backbone features.
This enriched visual representation leads to a clear performance gain over the earlier IO variants 
(Acc = 0.8305, F1 = 0.8218, AUC = 0.8972), 
demonstrating that incorporating region-level structural reasoning inside the image encoder improves robustness and multimodal alignment. 
However, the improvements remain moderate compared to our full BCA design, 
indicating that IO cues are most effective when combined with explicit interior--exterior contrast modeling and cross-modal interaction rather than being used solely within the visual stream.

\noindent\textbf{IO-Difference Enhanced Fusion.}
This variant further explores whether the contrast between the interior and exterior regions 
can serve as a reliable indicator of enclosure characteristics. 
Given the inner and outer visual tokens, we compute their similarity to assess whether the two regions exhibit meaningful structural discrepancy. 
When the similarity is low (below a threshold of 0.6), indicating a clear interior--exterior contrast that is typically observed in gated communities, 
we construct an IO-difference feature by subtracting the outer representation from the inner one and transforming it through an MLP. 
This difference vector acts as an additional contrast-aware cue and is fused together with the three modality-specific cross-attention outputs 
(image--text, image--numerical, and text--numerical). 
If the inner and outer regions are too similar (similarity above the threshold), 
the IO-difference signal is suppressed to avoid injecting unreliable enclosure cues.
While this conditional contrast modeling provides a reasonable improvement over naive IO gating 
(F1 = 0.8095, AUC = 0.8912), 
its performance still lags behind both the IO-free baseline and our full BCA formulation. 
This suggests that interior--exterior difference cues are indeed informative but must be contextualized through richer bi-directional cross-modal reasoning 
to fully capture the multi-faceted spatial semantics of gated and open communities.

\noindent\textbf{IO with BCA (ours).}
Our full formulation integrates both the interior--exterior difference feature $\bm{F}_{\text{diff}}$ 
and the numerical similarity statistics ($S_{\text{inn}}, S_{\text{out}}, S_{\text{img}}$), 
and processes them jointly inside a dedicated Bi-directional Cross-modal Attention (BCA) block. 
The BCA module injects these contrast-aware cues into the multimodal interaction pathway, 
producing discriminative IO features that are subsequently fused with the three modality-specific cross-attended features 
via the confidence-weighted fusion in Eqs.~(4)--(6). 
This design achieves the highest precision among all variants 
(Prec = 0.8737, a gain of $+5.68$ over Direct IO Fusion and $+1.96$ over Without IO), 
together with competitive overall metrics (Acc = 0.8311, F1 = 0.8184, AUC = 0.9025). 
Compared with the strongest naive IO baseline, BCA improves Acc by $+2.02$, F1 by $+1.15$, and AUC by $+1.33$ points.

Beyond accuracy, a key strength of our method lies in its stability: 
IO with BCA exhibits consistently smaller standard deviations across all five evaluation metrics 
(e.g., Acc $\pm$0.0227, Prec $\pm$0.0306, Rec $\pm$0.0318, F1 $\pm$0.0250, AUC $\pm$0.0257), 
which are notably lower than those of competing IO variants. 
The reduced variance demonstrates that BCA not only improves the mean performance but also produces 
more reliable cross-fold behavior and more robust multimodal alignment. 
Although recall shows a mild decrease compared with the IO-free model, 
the substantial precision gain and the pronounced reduction in variance make our formulation preferable for gated-community detection, 
where minimizing false positives and ensuring stable predictions are critical for downstream urban-analysis applications.

Overall, these ablations validate that IO cues are most beneficial when modeled through bi-directional attention and confidence-weighted multimodal fusion, rather than by simple gating or feature replacement.

\subsection*{D.4 Fixed Stratified 8:2 Splits by Area and Geographic Clustering}
\label{Supp:D4}

\noindent\textbf{Goal.}
To ensure reproducible benchmarking while balancing urban morphology and spatial context, we report results over five \emph{fixed} stratified 8:2 (train:test) splits for the Guangzhou evaluation benchmark. Each split is constructed by joint stratification over (i) AOI area and (ii) geographic clustering degree, together with the gated/open label.

\paragraph{Area stratification.}
For each AOI $i$ with polygon area $a_i$, we assign an area bin using quantile binning:
\begin{equation}
b^{(a)}_i=\mathrm{QuantileBin}(a_i;Q_a),
\label{eq:area_bin}
\end{equation}
where $Q_a$ is a set of quantile cut points (e.g., quartiles or deciles), ensuring proportional coverage of small to large compounds in every split.

\paragraph{Geographic clustering stratification.}
Let $\mathbf{c}_i\in\mathbb{R}^2$ denote the AOI centroid in a planar coordinate system. We partition AOIs into $K$ spatial clusters (approximating neighborhood-level geographic grouping) by clustering on centroids:
\begin{equation}
b^{(g)}_i=\mathrm{Cluster}(\mathbf{c}_i;K),
\label{eq:geo_cluster}
\end{equation}
where $\mathrm{Cluster}(\cdot)$ denotes a spatial clustering operator (e.g., $K$-means). This captures neighborhood-level geographic grouping and discourages splits dominated by a single sub-region.

\paragraph{Joint stratification and split construction.}
We define a joint stratum index:
\begin{equation}
s_i=\big(b^{(a)}_i,\;b^{(g)}_i,\;y_i\big),
\label{eq:joint_stratum}
\end{equation}
where $y_i\in\{0,1\}$ is the gated/open label. For each stratum $s$, we sample a fixed proportion $\rho=0.2$ into the test set and assign the rest to the training set:
\begin{equation}
\mathcal{D}^{(k)}_{\mathrm{test}}=\bigcup_{s}\mathrm{Sample}\!\left(\mathcal{D}_s,\rho;\,\text{split\_seed}_k\right),\quad
\mathcal{D}^{(k)}_{\mathrm{train}}=\mathcal{D}\setminus\mathcal{D}^{(k)}_{\mathrm{test}},\quad
k=1,\ldots,5.
\label{eq:stratified_split}
\end{equation}
Here $\mathcal{D}$ is the full labeled Guangzhou benchmark set, $\mathcal{D}_s=\{i\in\mathcal{D}: s_i=s\}$ is the subset in stratum $s$, and $\mathrm{Sample}(\cdot)$ samples without replacement within each stratum.

\paragraph{Validation split.}
For each split $k$, we construct a validation set by sampling 10\% of $\mathcal{D}^{(k)}_{\mathrm{train}}$ with the same joint stratification, and use it for early stopping and model selection.

\paragraph{Reporting and release.}
All quantitative results are reported as mean $\pm$ standard deviation over the five fixed splits.
To enable exact reproduction, we release the split indices as \texttt{splits/guangzhou\_split\_k.json} ($k=1..5$). In all experiments, we additionally fix a global training seed (\texttt{seed=42}) for Python/NumPy/PyTorch/CUDA (including dataloader worker seeds) to ensure identical sampling and initialization \emph{given the fixed split indices}.

\subsection*{D.5 Implementation Details.}
\label{Supp:D5}
All models are implemented in PyTorch. MCGC uses DINOv3-SAT as the visual backbone and applies LoRA for parameter-efficient fine-tuning. We train with AdamW (lr $5\times10^{-5}$, weight decay 0.01, batch size 64) for up to 30 epochs, using early stopping based on validation \textbf{AUC}. Unless otherwise specified, results are reported over five stratified runs with an 8:2 train/test split to preserve class balance. Training and inference are performed on four NVIDIA RTX 4090 GPUs.

\subsection*{D.6 Resolution Ablation}
\label{Supp:D6}
To test whether interior--exterior contrast is resolution-dependent, we downsample the remote-sensing input while keeping the same train/test protocol and non-visual modalities. The result confirms genuine sensitivity to image resolution: reducing input resolution from $224\times224$ to $112\times112$ lowers F1 from .849 to .740 and AUC from .917 to .880; further reducing to $56\times56$ lowers F1 to .672 and AUC to .765. Performance remains above chance even at $56\times56$, indicating that coarse urban-form cues remain useful, but high-resolution imagery is important for boundary, entrance, and edge-transition cues.

\begin{table}[!h]
\centering
\caption{Resolution ablation for remote-sensing imagery.}
\label{tab:resolution_ablation}
\begin{tabular}{lcc}
\toprule
\textbf{Input resolution} & \textbf{F1} & \textbf{AUC} \\
\midrule
$224\times224$ & .849 & .917 \\
$112\times112$ & .740 & .880 \\
$56\times56$ & .672 & .765 \\
\bottomrule
\end{tabular}
\end{table}

\subsection*{D.7 OSMnx Detour Check}
\label{Supp:D7}
The main detour simulation is geometry-based and intentionally does not use observed road networks because observed streets already encode historical routing around gated compounds. To check that the accessibility claim is not an artifact of this abstraction, we additionally construct an OpenStreetMap-based diagnostic using OSMnx v1.9. In this check, we obtain the local pedestrian network from OSM, remove or penalize paths crossing gated AOI barriers, and compare the shortest network path against the corresponding straight-line baseline. A representative gated-block example yields a detour index of DI=2.22, consistent with the geometry-based result that enclosed residential blocks can impose substantial pedestrian impedance. Because OSM completeness varies across Chinese cities, we treat this as a diagnostic validation rather than the primary benchmark protocol.

\begin{figure}[!h]
    \centering
    \includegraphics[width=0.75\linewidth]{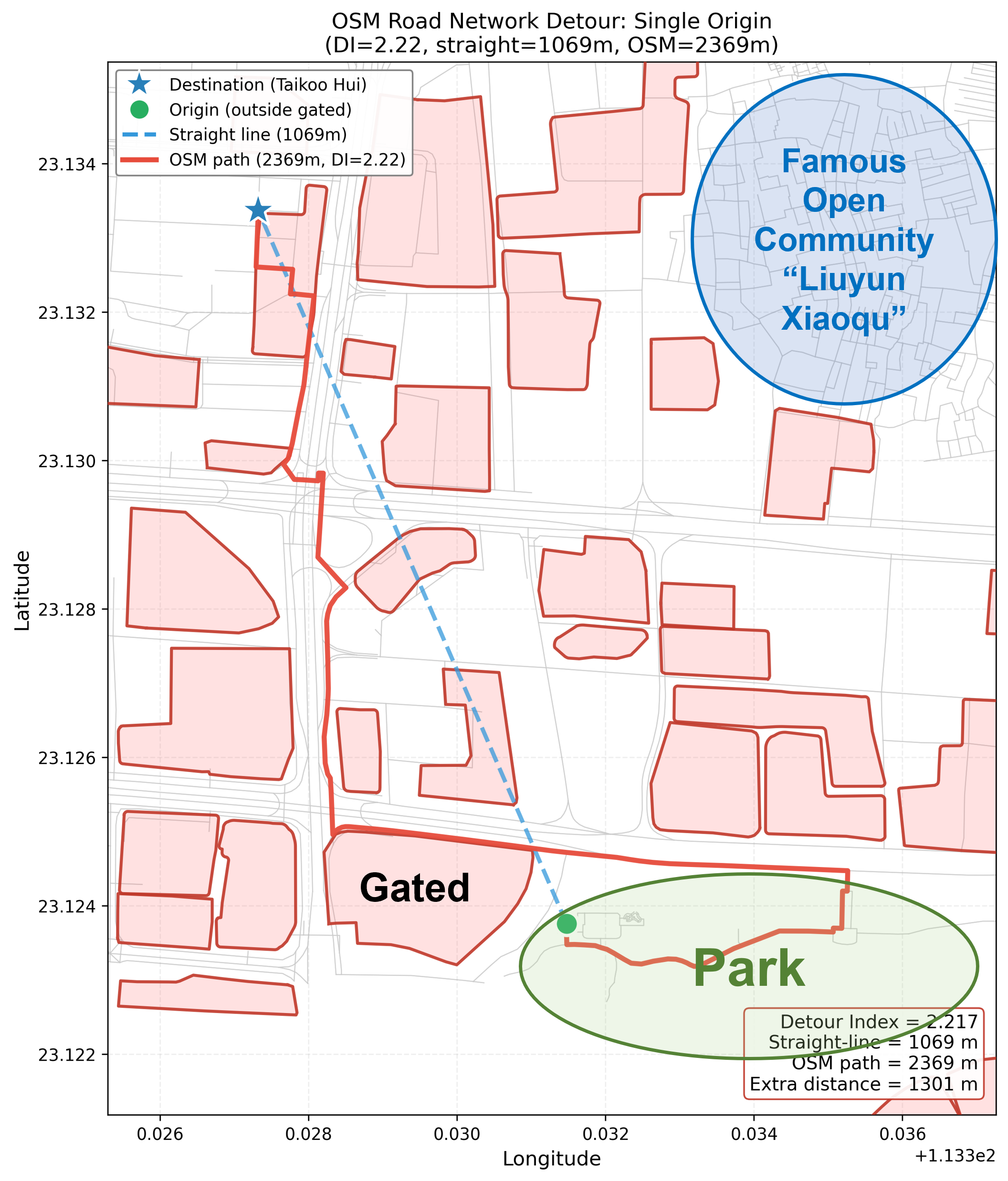}
    \caption{OSM+OSMnx diagnostic detour example. The example illustrates how a gated block can increase network travel relative to the straight-line baseline, yielding DI=2.22 in the rebuttal diagnostic.}
    \label{fig:osmnx_detour}
\end{figure}

\subsection*{D.8 Failure and Borderline Case Library}
\label{Supp:D8}
To make the operating regime of MCGC explicit, we construct a qualitative library of correct, borderline, false-positive, and false-negative cases. In the five-fold validation predictions, 139 samples fall into the low-confidence interval $p(\mathrm{gated})\in[0.45,0.55]$. Manual inspection shows that these cases concentrate in interpretable ambiguity types rather than random errors: partial gating, mixed-use villages, weak or occluded entrances, similar inside/outside textures, perimeter-like open compounds, and AOI polygon ambiguity.

\begin{figure}[!h]
    \centering
    \includegraphics[width=\linewidth]{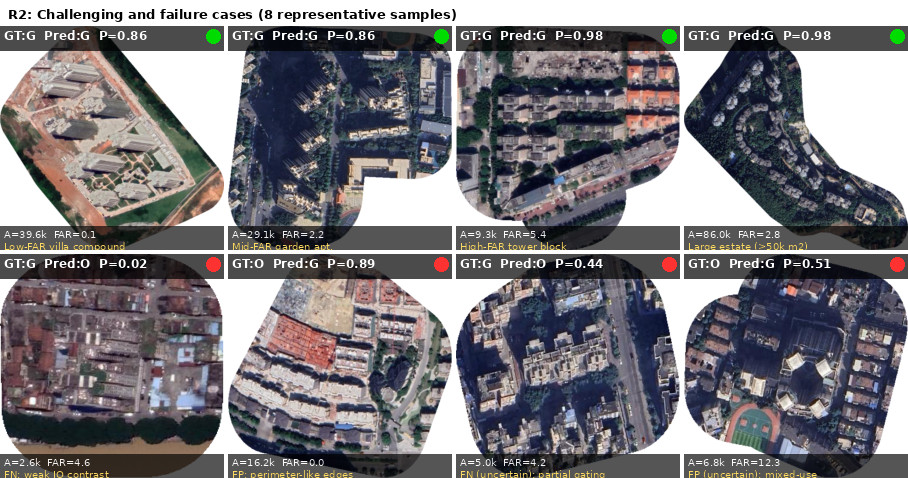}
    \caption{Representative correct, failure, and borderline cases. The gallery illustrates where IO contrast helps and where it remains insufficient: partial gating, mixed-use morphology, weak or occluded entrances, perimeter-like open compounds, and similar inside/outside textures.}
    \label{fig:failure_case_library}
\end{figure}

We organize the full visual case library as 104 cases across seven failure/ambiguity categories: (1) partial gating, (2) decorative or non-functional walls, (3) weak or occluded entrances, (4) mixed-use residential/commercial or village morphology, (5) AOI polygon ambiguity, (6) similar inside/outside textures, and (7) perimeter-like open compounds. For each category, we inspect false positives, false negatives, and low-confidence correct predictions across size and density strata. These cases show that MCGC relies on meaningful AOI localization and inside--outside contrast, but exact polygon shape alone is not the decisive signal; ambiguous enclosure remains a real labeling and modeling challenge. This is why we frame IO as a boundary-conditioned inductive bias, and leave polygon-free discovery, entrance detection, OSMnx connectivity, and local taxonomy refinement as future work.




\renewcommand{\thefigure}{E\arabic{figure}}
\renewcommand{\thetable}{E\arabic{table}}
\setcounter{figure}{0}
\setcounter{table}{0}
\section{Benchmark, Reproducibility, and Release Details}
\label{Supp:E}
\subsection{Benchmark Card and Reproducibility (Splits \& Seeds)}
\label{Supp:E.1}

\noindent\textbf{Benchmark card (GBA-GCs).}
\begin{table}[t]
\centering
\small
\setlength{\tabcolsep}{5pt}
\renewcommand{\arraystretch}{1.08}
\begin{tabular}{p{0.30\linewidth} p{0.64\linewidth}}
\toprule
\textbf{Item} & \textbf{Specification} \\
\midrule
Name & GBA-GCs (Greater Bay Area Gated Communities) \\
Task & AOI-level binary classification: \emph{gated} vs.\ \emph{open} \\
Scale & 37{,}444 residential AOIs (boundary polygons) \\
Region & Greater Bay Area (nine mainland cities); Guangzhou is used for controlled benchmark evaluation \\
Modalities & Satellite imagery (0.8\,m; 50\,m context buffer), Chinese metadata (name+address), structured attributes (e.g., FAR, POI statistics) \\
Labels & Expert-verified gated/open labels for the Guangzhou benchmark evaluation set; region-wide labels are obtained by model inference with post-hoc human checks \\
Reliability & Inter-annotator agreement reported with Cohen's $\kappa$ on a duplicated subset of N=\,200 AOIs (main paper Sec. 3.1) \\
Evaluation & F1 and AUC (main), reported as mean $\pm$ std across five fixed splits \\
Known limitation & Random splits may be affected by spatial autocorrelation; we mitigate this via joint stratification and report robustness checks (Supplementary Sec. D.4) \\
\bottomrule
\end{tabular}
\vspace{-4pt}
\end{table}

\noindent\textbf{Fixed splits and seeds.}
To ensure reproducibility, we use five \emph{fixed} stratified splits for Guangzhou evaluation:
\begin{itemize}
    \item \textbf{Split construction:} each split uses an 8:2 train/test ratio with joint stratification over AOI \textbf{area} and \textbf{geographic clustering} (together with the gated/open label) to balance urban morphology and geography (Supplementary Sec. D.4).
    \item \textbf{Validation:} for each split, we construct the validation set by taking 10\% of the training set with the same stratification, and use it for early stopping and model selection.
    \item \textbf{Split files:} we release split indices as \texttt{splits/guangzhou\_split\_k.json} ($k=1..5$); all results in the main paper are computed on these fixed splits.
    \item \textbf{Random seeds (training):} we fix a global training seed (\texttt{seed=42}) for Python/NumPy/PyTorch/CUDA for each run (including dataloader worker seeds) to guarantee identical sampling and initialization \emph{given the fixed split indices}.
    \item \textbf{Reporting:} all quantitative results are reported as mean $\pm$ std over the five fixed splits, unless stated otherwise.
\end{itemize}
\vspace{-4pt}

\noindent\textbf{Pointer to release/compliance details.}
The code, split files, release documentation, and benchmark tables are available at \dataseturl.

\subsection{Data Sources, Licensing, and Model Release}
\label{Supp:E.2}
\paragraph{Release structure.}
The official paper benchmark contains 37{,}444 residential AOIs and the experiments in this paper are tied to this fixed version. The public release is hosted at \dataseturl. It contains the paper benchmark documentation, code, model-interface examples, split/evaluation information, reconstruction workflows, dataset cards, and versioned public tables. We maintain the repository as a living research release: future GitHub versions may add non-sensitive AOI records, improved documentation, model checkpoints, and additional diagnostic tables, but the paper results should be interpreted against the 37{,}444-AOI benchmark described above.

\paragraph{Public anonymized tier.}
The public tier contains only fields that are intended for open research use: anonymous AOI identifiers, gated/open labels, WGS84 centroids and area for reconstruction, official split metadata where applicable, coarse or binned non-sensitive derived attributes, aggregate metadata, code, evaluation scripts, model cards, reconstruction workflows, and documentation. Public files do not redistribute community names, raw provider UIDs, addresses, raw AOI coordinates beyond centroids, AOI polygons/geometries, raw satellite or street-view imagery, or provider-owned metadata.

\paragraph{Controlled access tier.}
Researchers who need restricted components for exact reconstruction or audit can use the controlled-access portal linked from \dataseturl. The controlled repository contains the access policy, DUA template, request form, archive manifest, checksums, permitted/prohibited use rules, and small de-identified samples illustrating the data structure. Raw archives are not hosted on GitHub. Approved users receive access-controlled storage links only after review and signature of a non-commercial DUA.

\paragraph{DUA and prohibited uses.}
Controlled access is restricted to accountable, non-commercial research. The DUA prohibits redistribution, public posting of controlled files, attempts to re-identify locations beyond the approved research purpose, surveillance, geofencing, targeted enforcement, resident- or community-level profiling, commercial real-estate scoring, advertising/targeting, policing applications, and decisions affecting specific residents or communities. It also specifies storage controls, named-user access, deletion/return obligations, misuse reporting, and takedown procedures.

\paragraph{Reconstruction with licensed sources.}
Users who need spatial inputs must obtain map geometries, metadata, and imagery from the original providers under the corresponding terms of service. Our preprocessing scripts are provider-agnostic: AOIs and metadata may be reconstructed from Baidu, Amap, OSM, or other licensed local sources, and imagery may come from any licensed sub-meter source. Sentinel-2 at 10\,m is generally too coarse for boundary cues. We also document an OSM/GEE reconstruction path as an unrestricted open-source alternative when local OSM coverage is adequate; such reconstructions should be reported separately from the controlled benchmark because OSM completeness varies by region.

\paragraph{Third-party data compliance.}
Baidu/Amap AOIs and POIs, Google or other satellite imagery, and street-view evidence remain property of their respective providers and are not redistributed by us. The public release therefore focuses on anonymized labels, derived/binned attributes, model/code artifacts, and reproducibility documentation. External transfer or diagnostic sets are not part of the main benchmark contribution unless local permissions and definitions are separately established.

\paragraph{Model release.}
The trained MCGC checkpoint is released through the public repository's GitHub Releases with filename, checksum, model-card restrictions, and an example inference API. The visual remote-sensing modality is required at inference time, while text and numerical modalities may be missing through modality masks. The released model is intended for non-commercial research, benchmarking, and aggregate urban-science analysis only.

\paragraph{Licensing summary.}
\begin{itemize}
    \item \textbf{Public anonymized tables and documentation:} non-commercial research use with attribution and third-party-data disclaimer.
    \item \textbf{Controlled components:} non-commercial DUA only; no redistribution.
    \item \textbf{Models:} CC BY-NC 4.0 for research and non-commercial use.
    \item \textbf{Code:} MIT License, excluding third-party pretrained weights and data.
\end{itemize}
\paragraph{Disclaimer.}
The release is designed for aggregate research and benchmark reproduction. It should not be used for surveillance, targeting, policing, commercial profiling, or decisions affecting residents or specific communities.
